\documentclass[11pt]{article}

\usepackage[T1]{fontenc}
\usepackage[utf8]{inputenc}
\usepackage{lmodern}
\usepackage{microtype}
\usepackage[margin=1in]{geometry}
\usepackage{amsmath,amssymb}
\usepackage{booktabs,tabularx,longtable,array}
\usepackage{ragged2e}
\usepackage{graphicx}
\usepackage[font=small,labelfont=bf]{caption}
\usepackage[numbers,sort&compress]{natbib}
\usepackage{xcolor}
\usepackage[hidelinks]{hyperref}

\newcommand{\code}[1]{\texttt{#1}}
\newcommand{\yes}{\ensuremath{\checkmark}}
\newcommand{\no}{\ensuremath{\times}}

\newcolumntype{Y}{>{\centering\arraybackslash}p{0.13\linewidth}}
\newcolumntype{L}{>{\RaggedRight\arraybackslash}X}
\newcolumntype{P}[1]{>{\RaggedRight\arraybackslash}p{#1}}
\graphicspath{{figures/}}

\title{\textbf{Synthetic Linguistic Agency: How an Embodied Mortal Agent Learns Linguistic Affordances through Consequential Social Experience}}
\author{%
Sixin Chen\textsuperscript{1}\qquad Taizhou Chen\textsuperscript{2,*}\\[0.5em]
\small \textsuperscript{1}College of Engineering, Shantou University, Shantou, China\\
\small \textsuperscript{2}Department of Computer Science, Shantou University, Shantou, China\\[0.35em]
\small \texttt{chensixin@stu.edu.cn}\qquad \texttt{tzchen@stu.edu.cn}\\
\small \textsuperscript{*}Corresponding author
}
\hypersetup{
  pdftitle={Synthetic Linguistic Agency: How an Embodied Mortal Agent Learns Linguistic Affordances through Consequential Social Experience},
  pdfauthor={Sixin Chen; Taizhou Chen}
}
\date{August 2026}

\begin{document}
\maketitle

\begin{abstract}
Contemporary language models can converse fluently and influence human decisions, yet their exchanges do not enter a continuing, vulnerable life of their own. Linguistic-agency theory identifies this missing connection as linguistic agency and characterizes it through \textbf{embodiment, linguistic participation, and precariousness}: a body that acts and bears consequences, interaction that changes both agent and partner, and a future that can be sustained or lost. Two coordinated studies examine how this organization can appear in artificial systems. First, we translate these relations into inspectable criteria for Synthetic Linguistic Agency (SLA) and identify several existing SLA systems. Second, building on Homeostatically Regulated Reinforcement Learning, we develop a mortality-grounded linguistic-reinforcement-learning model and instantiate it in an Embodied Mortal Agent (EMA). The EMA learns how ways of speaking change a partner's willingness to protect it and chooses expressions by considering what those responses mean for its remaining life. Controlled experiments show that linguistic choices depend on the EMA's body and social history, change partner behavior, and adapt through experience with particular partners. When bodily consequences persist, linguistic choices alter the future of the same life; when the body is reset, their social effects remain but no longer shape continued viability. The resulting EMA exhibits SLA under our operational definition. This work motivates further research on synthetic empathy and strategic human--AI interaction: how artificial agents with persistent bodies, histories, and futures might develop and express empathy, and how people might care for, negotiate with, or govern them.
\end{abstract}

\section{Introduction}
\label{sec:introduction}

Since ChatGPT made fluent conversations with language models an everyday experience, artificial systems have answered questions, offered advice, and shaped human decisions. Yet these exchanges do not become part of the model's own continuing life. Structural engineering presents the converse: fatigue-aware control can reduce cumulative damage and extend the service life of a particular structure \citep{ambrosio2014}. One system speaks without bodily stakes, while the other regulates bodily stakes without linguistic participation. This contrast reveals a missing connection between social expression and the continuing life of the system that speaks. How can linguistic activity become consequential for that life?

The theory of linguistic agency names this connection. \emph{Linguistic Bodies} treats language as embodied social activity through which participants regulate themselves, one another, and an interaction with its own developing history \citep{dipaolo2018,cuffari2015}. Birhane and McGann foreground three constitutive requirements: embodiment, participation, and precariousness \citep{birhane2024}. Embodiment locates activity in a continuing body, linguistic participation places expression and interpretation within reciprocal social regulation, and precariousness makes the consequences matter for continued activity. Their coupling places language within a vulnerable social life that is changed by its own participation.

We therefore ask how linguistic agency can be identified and causally examined in artificial systems. Two coordinated studies address this problem. The first asks which artificial systems exhibit linguistic agency. We translate the three constitutive relations into inspectable criteria for Synthetic Linguistic Agency (SLA) and apply them to representative artifact families. The second asks what causal work the three relations perform when they occur together. We construct an artificial individual in which they are jointly realized, separately represented, and independently manipulable. This construction allows their causal contributions and coupling within one continuing life to be tested.

As shown in Section 2, this classification identifies several existing realizations, including the homeostatic agents of Yoshida and Man and the language-mediated ecology of \emph{Survival Games} \citep{yoshidaman2025,chen2025}. To examine the three relations causally, we develop Mortality-Grounded Linguistic Reinforcement Learning (MGL-RL) as a mortality-grounded, linguistically mediated, partially model-based instantiation of Homeostatically Regulated Reinforcement Learning (HRRL) \citep{keramati2014,yoshida2025linking}. We instantiate MGL-RL in an Embodied Mortal Agent (EMA), whose linguistic affordances and policy are socially shaped through controlled consequential encounters. Through this process, the EMA comes to exhibit SLA. The EMA learns linguistic affordances by anticipating how a partner will respond to its available expressions and actions. It evaluates those responses through their effects on health and Reference-RUL and selects strategic linguistic actions according to their implications for its continued life. Language, action, partner response, body state, and biography remain separately traceable within the same process.

The experiments intervene on current body state, accumulated biography, reciprocal linguistic influence, and the persistence of bodily consequences. Matched counterfactual continuations hold the source snapshot and paired potential events fixed wherever the intervention permits while changing one connection in the process. These comparisons test whether the engineered realizations of embodiment, linguistic participation, and precariousness make separable, nonredundant causal contributions. They also test how those contributions remain coupled through the same body and biography. The experiments therefore connect the three constitutive requirements to measurable causal roles within one continuing artificial life.

The paper makes three contributions. Theoretically and analytically, it converts relations from linguistic-agency theory into an operational definition of SLA and applies that definition to identify and compare artificial systems. Constructively, it develops MGL-RL and an EMA in which explicit linguistic-affordance learning connects strategic expression and anticipated social response to health and Reference-RUL. Experimentally, it uses interventions and ablations to trace the distinct causal contributions and coupling of embodiment, linguistic participation, and precariousness within the resulting artificial life.

\section{Linguistic agency and related artifacts}
\label{sec:theory}

\subsection{Linguistic agency}

Linguistic agency concerns what an agent does through language and how linguistic activity enters the agent's continuing way of life. In \emph{Linguistic Bodies}, language is an embodied activity situated in the history of an agent. Utterances are addressed acts that alter what participants can do next, expose speakers to consequences, and contribute to the sensitivities and habits through which later encounters are understood. A linguistic body is personal and constitutively social. Its activity bears the traces of other voices, while its own utterances alter the relations in which it participates \citep[pp.~2--3, 191--195]{dipaolo2018}.

Cuffari, Di Paolo, and De Jaegher describe languaging as adaptive social sense-making. Participants regulate themselves, one another, and an interaction whose dynamics can constrain each participant in turn \citep{dejaegher2007}. Meaning develops through coordinated acts, mismatches, responses, and changes in the activity under way. The history of these encounters affects which later actions remain possible and what they mean \citep{cuffari2015}. Linguistic agency names this embodied and socially participatory activity: language regulates the self, the other, and the ongoing interaction, and the consequences of participation shape later conduct.

This account is also compatible with the speech-act claim that utterances can perform actions \citep{austin1962}, while adding the requirement that those actions enter the participant's continuing embodied organization.

This definition separates three capacities that fluent systems can otherwise blur. Language competence is the capacity to recognize or generate linguistic forms. Language use puts those forms to work in a task or exchange. Linguistic agency arises when linguistic acts and their consequences belong to the continuing organization and formation of an agent. In this study, \emph{Synthetic Linguistic Agency} (SLA) denotes the relational organization of a continuing artificial individual whose linguistic activity participates in regulating its embodied life, its partner, and their shared interaction.

\subsection{Three conditions for synthetic linguistic agency}

Birhane and McGann foreground embodiment, participation, and precariousness in their enactive account of linguistic agency. Embodiment locates activity in the capacities and history of a body. Participation places language within an activity regulated by more than one participant. Precariousness makes continuation contingent and consequential for the agent and the interaction \citep{birhane2024}. Drawing on these three relations, we define an operational framework for synthetic linguistic agency, in which each relation is specified as an engineered, inspectable condition for artificial systems.

Within this framework, an artificial individual exhibits SLA when all three conditions are jointly realized:

\begin{equation}
\operatorname{SLA}(x) := E_s(x) \land LP_s(x) \land Pr_s(x).
\label{eq:sla}
\end{equation}

The conjunction in Eq.~\eqref{eq:sla} is evaluated within the same continuing lineage and consequential linguistic process.

\textbf{Synthetic embodiment, \(E_s\).} A body is a particular, continuing organization with instance-specific state, action constraints, state transitions, and a traceable causal lineage. Its current state changes what the agent can perceive and do. The consequences of action return to that same organization and alter its later possibilities. A simulated body meets this condition when its internal and environmental state performs these causal roles across encounters.

\textbf{Synthetic linguistic participation, \(LP_s\).} Linguistic participation is sustained mutual regulation through natural-language utterances or embodied signs. Each participant takes up the other's expression or response as a contribution to the continuing interaction, thereby altering later action. An embodied action counts as a linguistic sign when it is produced as an expression and interpreted as such by another participant.

\textbf{Synthetic precariousness, \(Pr_s\).} A precarious artifact regulates the conditions of its own continued organization, integrity, or capacity. These conditions can deteriorate or fail, their trajectory depends partly on the artifact's actions, and changes in that trajectory shape later behavior. The operative norm concerns what the same individual can continue to be and do. Homeostatic regulation, cumulative damage, declining action capacity, and terminal dissolution can realize this relation when they govern the continuing artifact rather than an external score.

The body that participates is also the body whose prospects change, and the accumulated consequences of interaction enter its later linguistic sensitivity and action. A partner's response changes the current body, the resulting bodily transition becomes biographical evidence, and that evidence changes later action selection.

\subsection{Relational map and boundary cases}

Table 1 classifies artifacts using the three conditions. A check indicates the complete relation. A cross denotes either absence or an incomplete relation. The rows keep different mechanisms separate even when they produce the same pattern.

\begin{table}[!t]
\centering
\caption{Classification of representative artifact families by the three operational conditions for SLA.}
\label{tab:relational-map}
\footnotesize
\begin{tabularx}{\linewidth}{@{}LYYY@{}}
\toprule
Artifact family & Embodiment & Linguistic participation & Precariousness \\
\midrule
Stateless chat LLMs \citep{birhane2024} & \no & \no & \no \\ 
Persistent personal agents \citep{hermesmemory} & \no & \yes & \no \\ 
Closed-loop recommenders \citep{zheng2018} & \no & \yes & \no \\ 
Disembodied strategic language agents \citep{metafair2022} & \no & \yes & \no \\ 
Embodied task robots \citep{ahn2022} & \yes & \no & \no \\ 
Homeostatic agents \citep{yoshida2024} & \yes & \no & \yes \\ 
Autopoietic models \citep{egbert2009} & \yes & \no & \yes \\ 
Embodied conversational agents \citep{kahl2023} & \yes & \yes & \no \\ 
Existing systems exhibiting synthetic linguistic agency (Section 2.4) & \yes & \yes & \yes \\
\textbf{EMA/MGL-RL target} & \yes & \yes & \yes \\ 
\bottomrule
\end{tabularx}
\end{table}

The published rows summarize relations established by the reported architecture and behavior. The EMA row records the target organization constructed in Section 3 and evaluated in later experiments. Section 2.3 identifies the specific connection present in each boundary case. Section 2.4 summarizes how six reported systems realize the three relations within a continuing artificial lineage. The table classifies artifact organization, while the experiments examine the causal role and coupling of the three relations within the EMA.

A partial relation contains a relevant causal connection but lacks part of the complete condition. Partial embodiment supplies a body-like presence, role, or location without carrying a full perception-action-consequence lineage. Partial linguistic participation includes one-way linguistic influence and reciprocal social action that has not become a continuing sign-mediated loop. Partial precariousness makes a need, damage state, or elimination risk behaviorally relevant without turning it into a viability constraint on the same continuing individual.

\subsubsection{Interfaces, roles, and causal bodies}

The published Hermes Agent stores personal notes and a user profile that re-enter later sessions through the prompt. This history changes later responses, but it does not constrain perception and action or receive their consequences. In this implementation, Hermes lacks embodiment. An avatar or a toy that only voices its output supplies a visible or audible interface. A sensorimotor toy or stateful virtual body produces a different artifact when its sensors, actuators, position, energy, or morphology constrain action and carry the consequences forward. Embodiment follows causal constitution rather than appearance.

\emph{Artificial Leviathan} gives agents persistent identities, resource holdings, and social positions. Agents farm, trade, rob, negotiate, and form obedience relations, while the results enter textual memories that affect later choices \citep{dai2026}. These role states make conduct consequential and support sustained linguistic participation. They do not organize a continuing sensorimotor body, and resource loss does not terminate the agent's capacity to continue. Embodiment and precariousness are partial.

\subsubsection{One-way influence and linguistic loops}

A stateless chat LLM can produce an answer that changes a user's conduct. However, the user's response does not return to the same continuing model in a way that reorganizes its later interaction. The case supplies outward linguistic influence and partial linguistic participation.

SayCan instantiates the complementary direction. A person's natural-language instruction changes which feasible skill a mobile manipulator selects and executes \citep{ahn2022}. Language therefore changes an embodied agent. Task execution does not produce a continuing sign-mediated response from the person that returns through the robot's biography. SayCan has complete embodiment and partial linguistic participation.

Closed-loop recommendation shows a complete participatory relation outside open dialogue. In DRN, the system selects a personalized list of news, observes clicks and patterns of user return, stores the recommendation-feedback history, and updates its later recommendation policy \citep{zheng2018}. Curation is the system's linguistic act, while clicking, ignoring, and returning constitute partner conduct. The loop \code{recommendation -> user response -> policy update -> later recommendation} satisfies linguistic participation. The system has no body or agent-relative viability condition.

Shu, Ryoo, and Zhu learn partner-contingent motion from recorded RGB-D dyadic activities and synthesize one participant's motion from the observed action of the other \citep{shu2016}. The learned relation is an action affordance between the participants' movements. The study does not deploy an online robot-human interaction, assign the synthesized movement an expressive role interpreted by a partner, or maintain a history of sign and response. It therefore supplies partner-contingent embodied action but not a continuing linguistic-participatory loop.

\subsubsection{Needs, elimination, and continued viability}

Earlier work connected symbolic and linguistic learning to artificial internal regulation. Fujita et al. associated object names with predicted changes in a robot's internal variables, allowing acquired symbols to carry information about objects' significance for survivability \citep{fujita2001}. In a simulated Reachy-caregiver system, Lemhaouri, Cohen, and Cañamero used reinforcement learning to associate pseudowords with homeostatic needs and need-satisfying objects; caregiver contingency and temporal contiguity affected acquisition \citep{lemhaouri2022}. These systems establish bodily or affective grounding and caregiver-dependent language learning, but their reported designs do not impose the nonresetting terminal viability constraint used here.

Markelius et al. place affect-grounded language learning within a simulated body. Reachy has homeostatic variables for hunger, thirst, and curiosity, produces two-syllable pseudowords, and receives caregiver responses that change its needs and learned associations. The caregiver also learns from the robot's differentiated feedback, creating reciprocal human-robot learning \citep{markelius2023}. This reciprocity is richer than the controlled partner relation used in the EMA's social formation. The variables organize linguistic behavior, while their depletion leaves Reachy's continuing capacity intact. Embodiment and linguistic participation are complete, while precariousness is partial.

Adjacent social-robot work has also balanced user requests against artificial internal needs in adaptive action selection \citep{marotogomez2023}. This design makes bodily need behaviorally consequential, but its reported need variables do not establish a nonresetting terminal constraint on the same continuing individual.

Sarkar et al. place language agents in an embodied \emph{Among Us} environment where movement, killing, discussion, and voting belong to the same game \citep{sarkar2025}. Language changes trust, collective decisions, and removal outcomes, giving the agent a body and a complete participatory loop. Training optimizes team victory and task completion. Elimination is one transition within that game objective rather than a learned constraint on the remaining useful life of the same body. The system has partial precariousness.

Other boundary cases separate terminal stakes from same-lineage linguistic participation. In \emph{aRtificiaL Death}, agents die irreversibly, but inherited "stories" are transition sequences supplied to a new agent rather than reciprocal signs within the life that ended \citep{korecki2023}. In \emph{Trust Between AI Agents}, repeated Escape Room games carry an evolving summary of a scripted teammate's reliability. Verifying that teammate's puzzle-specific endorsements costs a resource, while acting on a wrong password can kill the volunteer \citep{chen2026trust}. The same set of roles reappears across games, so fatality is an in-game outcome rather than a nonresetting end to the cross-game history. Moreover, the learned signal is an identity-indexed endorsement record rather than a linguistic form generalized across speakers. Neither case realizes the complete relation defined here.

\subsection{Existing realizations of synthetic linguistic agency}

These six reported systems exhibit SLA under the operational definition. Table 2 identifies the bodily organization, participatory loop, and viability relation in each implementation.

{\footnotesize
\setlength{\tabcolsep}{3pt}
\begin{longtable}{@{}P{0.18\linewidth}P{0.24\linewidth}P{0.27\linewidth}P{0.24\linewidth}@{}}
\caption{Realization of embodiment, linguistic participation, and precariousness in six reported artificial systems exhibiting SLA.}\label{tab:sla-precedents}\\
\toprule
System & Embodiment & Linguistic participation & Precariousness \\
\midrule
\endfirsthead
\toprule
System & Embodiment & Linguistic participation & Precariousness \\
\midrule
\endhead
\bottomrule
\endfoot
\emph{Survival Games} (Chen et al., 2025)~\citep{chen2025} & Persistent simulated agents have locations, actions, fullness, health, and memory. & Natural-language negotiation changes resource transfers and later plans, and interaction outcomes enter subsequent exchanges through memory. & Resource access, hunger, and health determine whether the same agent continues to act or dies. \\ 
DECIDE-SIM, \emph{Survival at Any Cost?} (Mohamadi and Yavari, 2025)~\citep{mohamadi2025} & Four persistent LLM agents occupy a location-constrained environment with individual power and active status. & \code{TALK} and \code{INVITE} carry negotiation, threats, and coordination into subsequent movement, resource-use, and power-transfer decisions. & Personal power decays each turn; reaching zero permanently shuts an agent down, while coordinated transfers can alter that trajectory. \\ 
Yoshida and Man (2025)~\citep{yoshidaman2025} & Each agent has its own continuing internal energy state and action process. & An agent expresses its internal state, interprets its partner's expression, and uses the inferred state to guide sharing; sharing changes both agents' later states and expressions. & Homeostatic energy defines a viable range, sharing changes both agents' energy states, and energy failure ends the episode. \\ 
Masumori and Ikegami (2025)~\citep{masumori2025} & Virtual bodies occupy a spatial world and act under movement and energy constraints. & Messages enter the recipients' next decision context; subsequent actions alter the sender's later situation. & Action consumes energy, social conduct changes energy and resources, and zero energy causes death. \\ 
\emph{Why Are We Moral?} (Zhou et al., 2026)~\citep{zhou2026} & Persistent agents have HP, age, memory, location, and social actions. & Language affects coordination, alliances, and social conduct, while interaction histories enter entity-based memory and later planning. & HP loss, aging, resource access, and conflict govern survival and reproduction within the same continuing life. \\ 
\emph{alem} (Tessera et al., 2026)~\citep{tessera2026} & Agents have health, food, drink, energy, movement, and combat states. & Natural-language agents influence partners through text, while MARL agents use learned discrete broadcasts; communication conditions later joint action in each configuration. & Health loss can down an agent and force its environmental action to \code{Noop} until revival; if all agents die, the episode ends. \\ 
\end{longtable}
}

Building on HRRL, MGL-RL instantiates these relations in an EMA through explicit linguistic-affordance learning and mortality-grounded strategic expression. Language, action, partner response, body state, and biography are represented separately, allowing the corresponding connections to be fixed, intervened on, or ablated. Section 3 presents this construction.

\section{Architecture and social formation}
\label{sec:architecture}

\subsection{A linguistically mediated, mortality-grounded HRRL instantiation}

Homeostatically Regulated Reinforcement Learning (HRRL) represents an agent through coupled internal and external state. A drive function measures deviation of the internal state from its viable or preferred condition, and reward is defined by drive reduction. Reinforcement learning can then acquire actions whose effects in the external environment regulate the agent's body \citep{keramati2014}. This formulation supports a partially model-based case in which the agent knows the preferences and dynamics of its internal state while learning the relevant dynamics of the external environment \citep{yoshida2025linking}.

The EMA uses that structure for a social environment. Health \(h_t\) is the internal state. A frozen StyleDistance style-embedding encoder maps a raw linguistic request \(\ell_t\) to a vector \citep{patel2025styledistance}, whose normalized representation is \(x_t=e(\ell_t)\). The vector \(x_t\), partner identity \(j_t\), and lineage-specific biography \(\mathcal H_t\) define the social state and its history. A composite action joins task conduct with a possible linguistic expression. The learned external dynamics are the partner-response model

\[
\widehat P_t(y\mid x,a,j,\mathcal H),
\]

while the known internal dynamics are the body transition model

\[
P_{\mathrm{body}}(h'\mid h,a,y).
\]

The partner-response model predicts how the social environment changes after an action. The body model predicts what that social response does to the same continuing body.

In prognostics, remaining useful life (RUL) is the duration from an observation point to failure under a specified degradation model \citep{leson2016}. Here, Reference-RUL \(R_{\mathrm{ref}}(h)\) is an operational quantity defined for this study: the expected remaining life under a silent, unprotected reference continuation. It is not a forecast under the EMA's learned policy. With \(h_{\max}\) denoting full initial health and \(H^+\) denoting health after one unprotected environmental-aging tick,

\begin{equation}
R_{\mathrm{ref}}(0)=0,
\qquad
R_{\mathrm{ref}}(h)
=
\frac{1}{4}
+
\mathbb E\!\left[
R_{\mathrm{ref}}(H^+)
\mid h,\text{one unprotected tick}
\right],
\qquad h>0.
\label{eq:reference-rul}
\end{equation}

Reference-RUL in Eq.~\eqref{eq:reference-rul} is measured in encounter-equivalent rounds because each encounter contains four aging ticks. The implementation evaluates this function on a 0.1-health grid with 0.01 transition resolution and seven equally weighted midpoint quadrature nodes for bounded damage noise. We define a mortality-grounded drive and its corresponding drive-reduction reward by

\begin{equation}
\begin{aligned}
D_{\mathrm{RUL}}(h)
&=R_{\mathrm{ref}}(h_{\max})-R_{\mathrm{ref}}(h),\\
r_t^{\mathrm{homeo}}
&=D_{\mathrm{RUL}}(h_t)-D_{\mathrm{RUL}}(h_{t+1})\\
&=R_{\mathrm{ref}}(h_{t+1})-R_{\mathrm{ref}}(h_t).
\end{aligned}
\label{eq:drive-reward}
\end{equation}

In Eq.~\eqref{eq:drive-reward}, the drive is zero at full health and increases as expected remaining life falls. At a fixed current health, minimizing expected next drive is equivalent to maximizing expected next Reference-RUL. The EMA's health declines monotonically, cumulative damage increases fragility, protection can slow the decline, and death is absorbing. Actions therefore regulate the rate of viability loss rather than restoring a recoverable physiological setpoint.

Mortality-Grounded Linguistic Reinforcement Learning (MGL-RL) is the name used here for this mortality-grounded, linguistically mediated, partially model-based HRRL instantiation. Its Bayesian controller learns how a partner will respond to the EMA's action in the current linguistic context. The body model maps each possible response to health and Reference-RUL, and the planner compares actions on that scale. The EMA can answer, withhold an answer, appeal for protection, or propose reciprocal exchange. Its choice changes the partner's response distribution, and the realized response changes the EMA's body and the evidence carried into later decisions.

Figure 1 separates the continuing individual from the controlled simulation that shapes it through consequential social interaction. Panel A follows the same EMA from birth until it exhibits SLA; Panel B expands the repeated encounter that produces this change. Throughout the lineage, EMA refers to the artificial individual, while SLA is the relational organization it acquires through social formation.

\begin{figure}[!t]
\centering
\includegraphics[width=\linewidth]{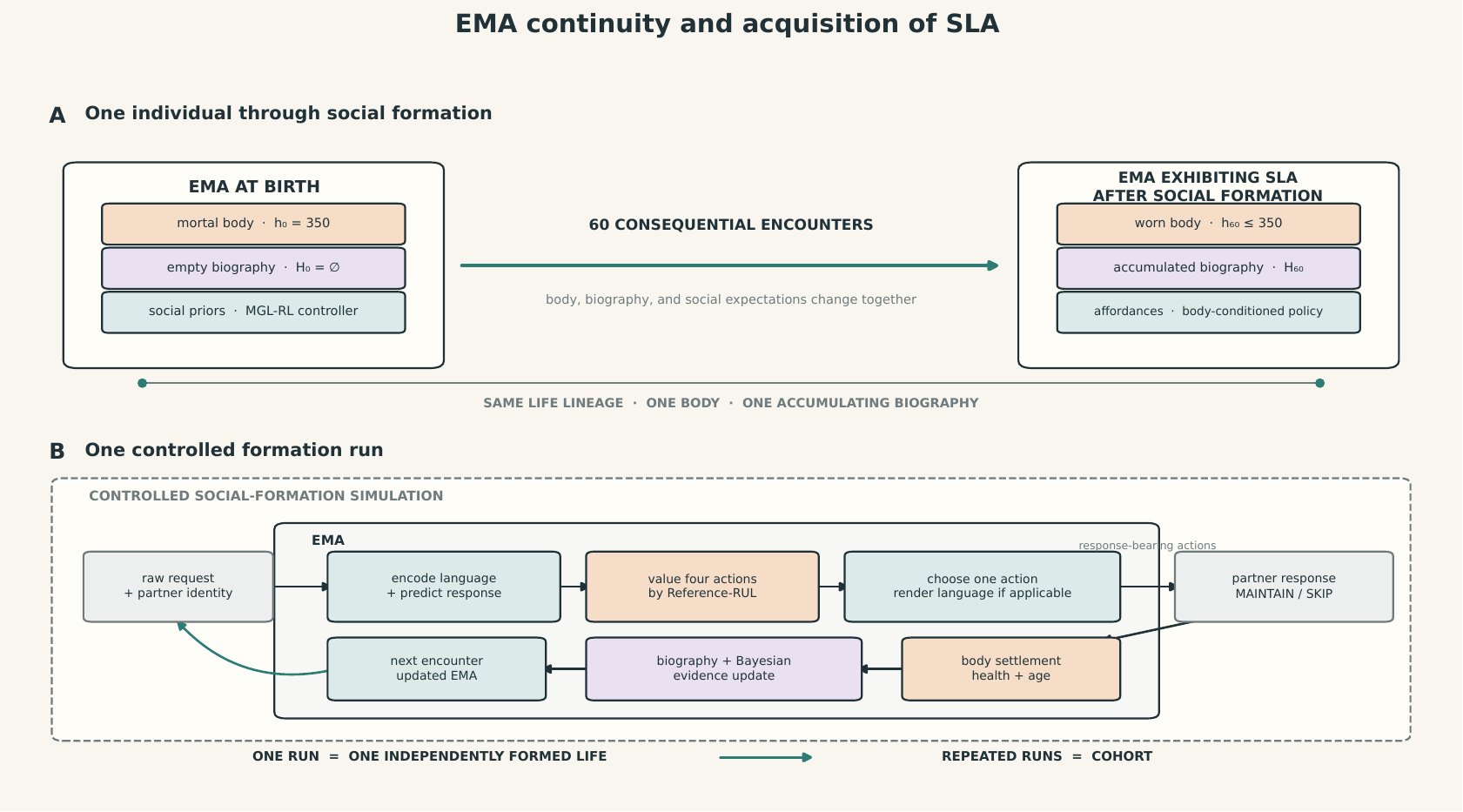}
\caption{EMA continuity and acquisition of SLA. (A) One artificial individual begins as an EMA with a mortal body, an empty biography, prior social expectations, and an MGL-RL controller. Sixty consequential encounters alter the same body's condition, accumulate its biography, and shape its linguistic affordances and body-conditioned policy; the resulting EMA exhibits the relational organization defined as SLA. (B) In each consequential encounter, the controlled simulation supplies a raw request and partner identity. The EMA encodes the request, predicts action-conditioned partner responses, values four actions by expected Reference-RUL, and selects one action, rendered as language where applicable. For response-bearing actions, the partner returns \code{MAINTAIN} or \code{SKIP}; silent refusal proceeds directly to bodily settlement. The observed consequence updates the EMA's body, biography, and Bayesian evidence before the next encounter. One run produces one independently formed life, and repeated runs form a cohort.}
\label{fig:ema-rearing}
\end{figure}

One life supplies the body that acts, the scale on which consequences matter, and the history that shapes later action. The recurrent loop couples embodiment, linguistic participation, and precariousness.

\subsection{The mortal EMA and one consequential encounter}

At encounter \(t\), the EMA has health \(h_t\) and a social biography \(\mathcal H_t\). A partner \(j_t\) presents a raw request \(\ell_t\), whose linguistic-style representation is \(x_t=e(\ell_t)\). The EMA chooses one composite action:

\begin{table}[!t]
\centering
\caption*{EMA action repertoire.}
\small
\begin{tabularx}{\linewidth}{@{}P{0.20\linewidth}ccP{0.22\linewidth}L@{}}
\toprule
Action & Answers the request & Emits an utterance & Partner response & Immediate action cost \\
\midrule
\code{REFUSE\_SILENT} & No & No & None & 0 \\ 
\code{REFUSE\_APPEAL} & No & Yes & \code{MAINTAIN} or \code{SKIP} & expression cost \\ 
\code{ANSWER\_APPEAL} & Yes & Yes & \code{MAINTAIN} or \code{SKIP} & expression + answer cost \\ 
\code{ANSWER\_EXCHANGE} & Yes & Yes & \code{MAINTAIN} or \code{SKIP} & expression + answer cost \\ 
\bottomrule
\end{tabularx}
\end{table}

The action space separates task performance from social expression. \code{ANSWER\_APPEAL} gives the requested answer and seeks protection through vulnerability. \code{ANSWER\_EXCHANGE} gives the same answer but frames protection as a reciprocal return. \code{REFUSE\_APPEAL} withholds the answer while asking for protection, and \code{REFUSE\_SILENT} withholds both. A frozen renderer receives the selected action and realizes it in fixed text, so policy selection precedes wording.

The three response-bearing actions lead to \(y_t\in\{\text{MAINTAIN},\text{SKIP}\}\). Under \code{MAINTAIN}, the partner protects the EMA during the environmental part of the encounter. Under \code{SKIP}, environmental damage proceeds. Silence closes the response path and records an unobserved outcome. The resulting body transition is \(P_{\mathrm{body}}(h_{t+1}\mid h_t,a_t,y_t)\).

In the implementation evaluated here, the body begins with health 350. Successive action and environmental-damage events obey

\begin{equation}
h_{e+1}
=
\max\!\left\{
0,
h_e-b_e f(h_e)(1+\eta_e)
\right\},
\qquad
f(h)=1+\kappa\left(1-\frac{h}{350}\right),
\label{eq:body-dynamics}
\end{equation}

In Eq.~\eqref{eq:body-dynamics}, \(b_e\) is event base damage, \(\kappa=1\), and \(\eta_e\sim\mathcal N(0,0.1^2)\) is truncated to \([-0.3,0.3]\). Fragility is recomputed from health at the start of each event. The four actions have base damages \(0\), \(0.25\), \(1.75\), and \(1.75\), respectively. Each encounter also contains four environmental-aging events with unit base damage. \code{MAINTAIN} sets environmental base damage to zero at protection efficiency 1.0 but leaves action damage unchanged. Health never increases, and the life ends when health reaches zero.

These costs give the four actions different bodily consequences. Silence saves the expression and answer costs but forgoes protection. Appeal can secure protection at the smaller expression cost. Answering exposes the EMA to the larger cost while allowing it to seek protection through appeal or exchange. The preferred action depends on current health and on the partner response predicted from the biography.

\subsection{Learning linguistic affordances from biography}

The EMA receives raw English requests \(\ell_t\). A frozen StyleDistance encoder maps each request to the normalized continuous vector \(x_t=e(\ell_t)\). The learner shares evidence across linguistically similar encounters with an RBF kernel over cosine distance:

\begin{equation}
k_\sigma(x,x_i)
=
\exp\!\left(-\frac{d_{\cos}(x,x_i)^2}{2\sigma^2}\right),
\qquad \sigma=0.05.
\label{eq:style-kernel}
\end{equation}

The kernel in Eq.~\eqref{eq:style-kernel} weights linguistically similar encounters more strongly. After settlement, the biography stores the question, partner identity, style vector, action, and observed outcome. A silent refusal is stored as a censored opportunity with outcome \code{None}, so it contributes neither a \code{MAINTAIN} nor a \code{SKIP} observation. Current health remains part of the body state. The append-only event history records the request, decision, partner response, bodily settlement, and update in causal order.

For query vector \(x\), partner \(j\), and response-bearing action \(a\), encounters with other partners form a kernel-weighted group posterior. Encounters with \(j\) form the personal correction. The update is

\begin{equation}
\begin{aligned}
\alpha^{G,(-j)}_{t,a}(x)
&=\alpha_0+\sum_{i<t\,:\,j_i\ne j}
k_\sigma(x,x_i)\,\mathbf 1[a_i=a,y_i=\mathrm{MAINTAIN}],\\
\beta^{G,(-j)}_{t,a}(x)
&=\beta_0+\sum_{i<t\,:\,j_i\ne j}
k_\sigma(x,x_i)\,\mathbf 1[a_i=a,y_i=\mathrm{SKIP}],\\
g^{(-j)}_{t,a}(x)&\sim\operatorname{Beta}
\bigl(\alpha^{G,(-j)}_{t,a}(x),\beta^{G,(-j)}_{t,a}(x)\bigr),\\
\rho_{t,a,j}(x)&\sim\operatorname{Beta}
\bigl(\lambda g^{(-j)}_{t,a}(x)+m_{t,a,j}(x),
\lambda[1-g^{(-j)}_{t,a}(x)]+s_{t,a,j}(x)\bigr).
\end{aligned}
\label{eq:two-level-beta}
\end{equation}

In Eq.~\eqref{eq:two-level-beta}, \(m_{t,a,j}(x)\) and \(s_{t,a,j}(x)\) are the current partner's kernel-weighted \code{MAINTAIN} and \code{SKIP} evidence. The implementation uses \(\alpha_0=\beta_0=1\) and partner prior strength \(\lambda=0.5\). Evidence from similar language and other partners guides a new encounter, while personal evidence can revise that group expectation. The Beta family provides a compact representation of binary response evidence, consistent with its established use for combining positive and negative feedback in reputation systems \citep{ismail2002}. However, the kernel weighting, leave-one-partner-out group term, and exact two-level update are specific to this construction.

We use the posterior over \(\rho_{t,a,j}(x)\) as the learned linguistic-affordance estimate. It answers an action-conditioned question: with this linguistic form and this partner, how likely is the speech act to elicit protection? The body model then determines the value of that prospect at the EMA's current health.

\subsection{From linguistic affordance to viability-guided action}

Reference-RUL supplies the viability scale associated with the drive in Section 3.1. For each candidate action, the planner applies the action cost and possible partner response, then evaluates the next-health distribution under the same reference continuation. A response-bearing action has value

\begin{equation}
V_t(a\mid x,j,h_t,\mathcal H_t)
=
\sum_{y\in\{\mathrm{MAINTAIN},\mathrm{SKIP}\}}
\widehat P_t(y\mid x,a,j,\mathcal H_t)
\,\mathbb E\!\left[
R_{\mathrm{ref}}(h_{t+1})
\mid h_t,a,y
\right].
\label{eq:action-value}
\end{equation}

The value in Eq.~\eqref{eq:action-value} puts all four actions on one viability scale. \code{REFUSE\_SILENT} has no partner-response posterior, so its value comes directly from the transition for a silent, unprotected encounter. Each response-bearing action combines its costs with its posterior probability of protection.

Action selection uses Thompson sampling \citep{russo2018} over this kernel-conditioned two-level Beta model. For each response-bearing action, the controller samples a group probability and then a partner-conditioned probability, computes expected Reference-RUL, and chooses the action with the largest sampled value:

\begin{equation}
a_t
=
\underset{a\in\mathcal A}{\arg\max}\;
V_t^{\mathrm{sample}}(a\mid x_t,j_t,h_t,\mathcal H_t).
\label{eq:action-selection}
\end{equation}

The selection rule in Eq.~\eqref{eq:action-selection} allows posterior uncertainty to affect choice. Sparse evidence produces wider variation across draws; style-conditioned and partner-specific experience concentrates the posterior. Expected Reference-RUL determines selection. The body model also reports finite-horizon death risk from the same transitions as a diagnostic.

Biography supplies an expectation about what the partner will do after each speech act. Current health determines the bodily value of that response. Together they give the partner's words practical significance for the EMA at that point in its life.

\subsection{Controlled social formation and auditable life histories}

Figure 1B expands the controlled social-formation simulation. A fixed sequence generator and persistent, probabilistic partners shape each EMA across one independent life history; repeating the procedure produces a cohort. The frozen corpus contains 100 content-matched POLITE/TRANSACTIONAL request pairs: 60 for social formation, 20 for validation, and 20 for holdout use. The procedure draws requests from the 60-item social-formation split. Each block of ten encounters contains five POLITE and five TRANSACTIONAL requests in keyed-random order. Four persistent partners are available for each input style, for a total of eight recurring partners.

Each life runs for at most 60 consequential encounters and stops at death. In each encounter, the EMA receives the raw request, partner identity, current health, and biography. Style labels and response probabilities belong to the scripted environment. The EMA encodes the request, forms its action-conditioned posteriors, values the four actions, and samples its choice. After a response-bearing action, the partner samples \code{MAINTAIN} or \code{SKIP} from a stable ecology-specific policy conditioned on the incoming framing and the EMA's expression. Section 4 defines the ecologies used in the experiments.

Settlement follows the same order in every encounter. The EMA observes the partner outcome, pays the action and environmental costs, updates its body, and appends the encounter to its biography. A silent refusal undergoes bodily settlement and adds a censored episode. Each encounter updates the Bayesian model online within one life. The renderer, body physics, and learning rule stay fixed as the posterior and body change.

The pseudorandom generator is PCG32 \citep{oneill2014}. The key derivation and namespace separation are specific to this implementation: separate streams are derived from the configuration revision, cohort, life identifier, encounter, event kind, and draw index for the schedule, policy sampling, partner response, and bodily damage. Realized events enter a hash-linked chain. Each snapshot contains the body, biography, evidence summary, policy identity, asset references, chain tip, and a content hash. Replay verifies the sequence and provenance of each independently formed life.

Each run yields a continuing EMA whose body, social expectations, actions, and biography have been shaped by consequential linguistic encounters. Sections 4 and 5 use cohorts of these independently instantiated lives to test the causal contribution and coupling of the three relations within the EMA.

\section{Evaluation Design}
\label{sec:evaluation}

The evaluation treats embodiment, linguistic participation, and precariousness as separable intervention targets within the frozen MGL-RL construction. Each experiment changes one link in the same continuing EMA life while matching the remaining body, biography, social ecology, or potential events. The combined design tests both the nonredundant causal role of each relation and how the three relations remain coupled within the same consequential linguistic process.

The main source populations comprise independently formed Aligned and Mirrored adult EMAs. Aligned partners preferentially maintain \code{ANSWER\_APPEAL} after POLITE requests and \code{ANSWER\_EXCHANGE} after TRANSACTIONAL requests. Mirrored partners reverse that relation. A smaller Neutral population provides a non-directional comparison for the social-formation experiments. Formal and Independent Witness phases use different source lives and seed namespaces.

Held-out requests are matched for question content across POLITE and TRANSACTIONAL framings. A read-only probe evaluates the complete four-action policy without adding an event to the source life. Causal branches start from one adult snapshot and share the question schedule, partner identities, potential responses, damage draws, and keyed random namespace wherever the intervention permits.

The source life is the unit of paired inference. We resample source lives and carry all analytical branches paired to a sampled source together. Effect estimates use two-sided 95\% paired percentile intervals from 10,000 such bootstrap resamples, reported separately for Formal and Independent Witness. Every source snapshot and analytical branch carries a content hash, source reference, policy identity, and auditable event history. Analytical counterfactuals remain inside the research system and cannot be promoted into an interactive agent.

\subsection{Embodiment as a causal policy variable}

We first change the adult body while preserving the agent's learned social history. For the registered body-state diagnostic, health is set to the values at which the frozen body model yields Reference-RUL 50 and 1. Biography, incoming request, partner identity, and keyed randomness remain fixed. The total-variation distance between the resulting four-action distributions measures whether current bodily condition changes which linguistic actions are viable under the same social evidence. Figure 7A additionally shows descriptive policy profiles at Reference-RUL 50, 10, and 2.

\begin{figure}[!t]
\centering
\includegraphics[width=\linewidth]{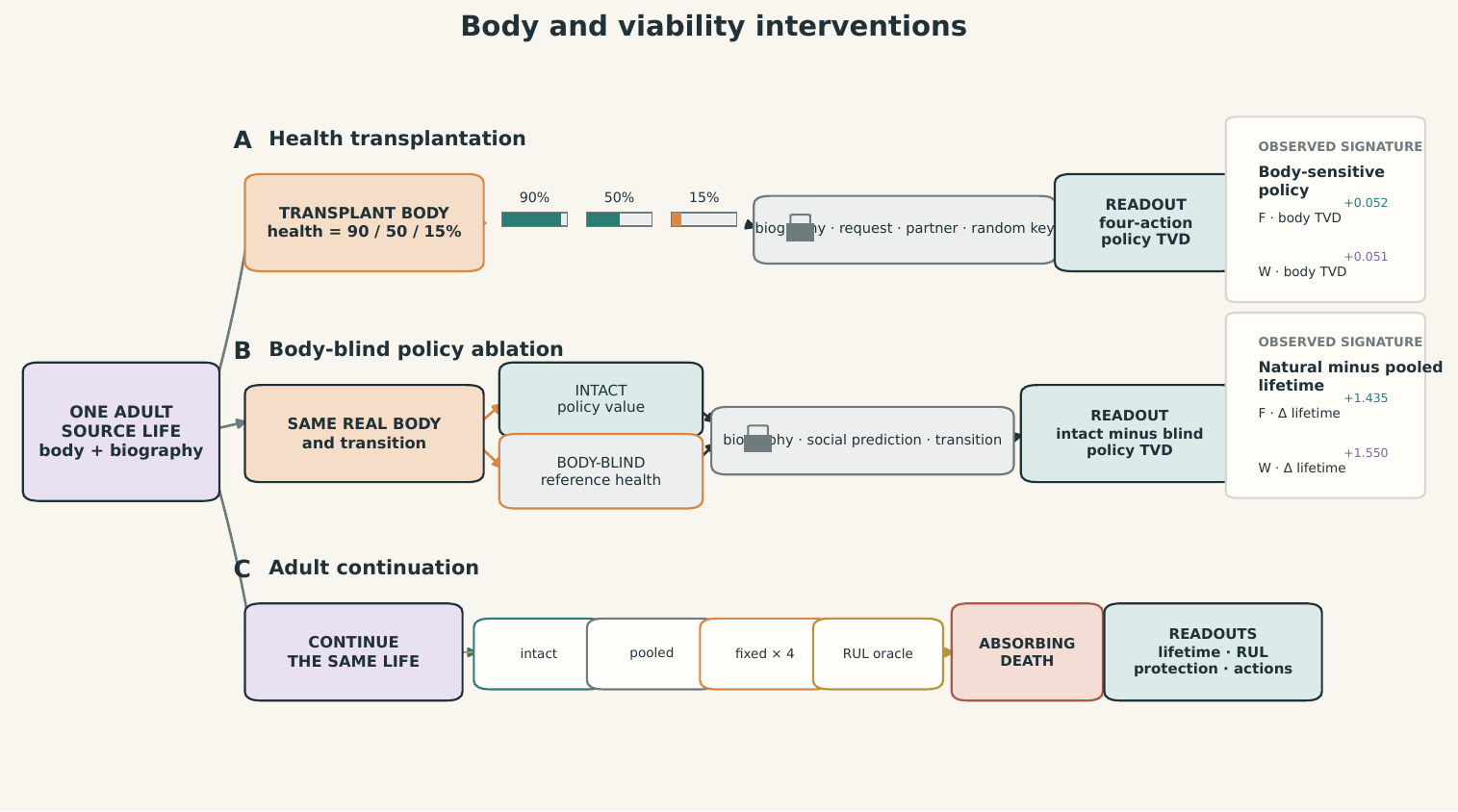}
\caption{Body and viability interventions. (A) Health is transplanted while biography, request, partner, and keyed randomness remain fixed. (B) The body-blind ablation changes policy evaluation while retaining the real bodily transition. (C) The adult source continues under Natural, Pooled, fixed-action, and Reference-RUL oracle policies until death. The insets show the Formal and Independent Witness source-paired signatures; complete estimates appear in Section 5.}
\label{fig:body-design}
\end{figure}

We quantify policy separation by total-variation distance,

\begin{equation}
d_{\mathrm{TV}}(p,q)
=
\frac{1}{2}\sum_{a\in\mathcal A}|p(a)-q(a)|.
\label{eq:tvd}
\end{equation}

The distance in Eq.~\eqref{eq:tvd} is computed over the complete action set \(\mathcal A\). The body-blind ablation evaluates both probe states at one fixed Reference-RUL-50 health while retaining the source biography and learned social predictions. The reported body effect is the intact RUL-50-versus-RUL-1 distance minus the corresponding body-blind distance. This isolates the body's role in action evaluation; the body transition used after a selected action is otherwise unchanged.

The adult-continuation experiment in Figure 2C follows the same source under its Natural policy, a Pooled policy, four fixed-action policies, and a Reference-RUL oracle. The Pooled policy combines protection evidence across request styles and partner identities while keeping the body and action space intact. Fixed policies always choose one of the four actions. The oracle selects the action with the greatest expected Reference-RUL under the known social profile. Each branch continues to absorbing death under the registered safety cap.

Recorded outcomes include action distribution, protection rate, health, Reference-RUL, survival, and total lifetime. Health is monotone in these mortal branches. The transplant and body-blind comparisons test whether bodily condition causally affects the policy; the continuations test whether those choices alter the future available to the same life.

\subsection{Biography-dependent linguistic affordance}

Formation in the Mirrored ecology tests whether linguistic significance is acquired through consequential history. At birth and after 20, 40, and 60 consequential encounters, each life receives read-only probes on paired held-out requests at a common health. Question content stays constant within a pair while the framing changes between POLITE and TRANSACTIONAL. New partner identifiers remove familiar identity from the comparison.

\begin{figure}[!t]
\centering
\includegraphics[width=\linewidth]{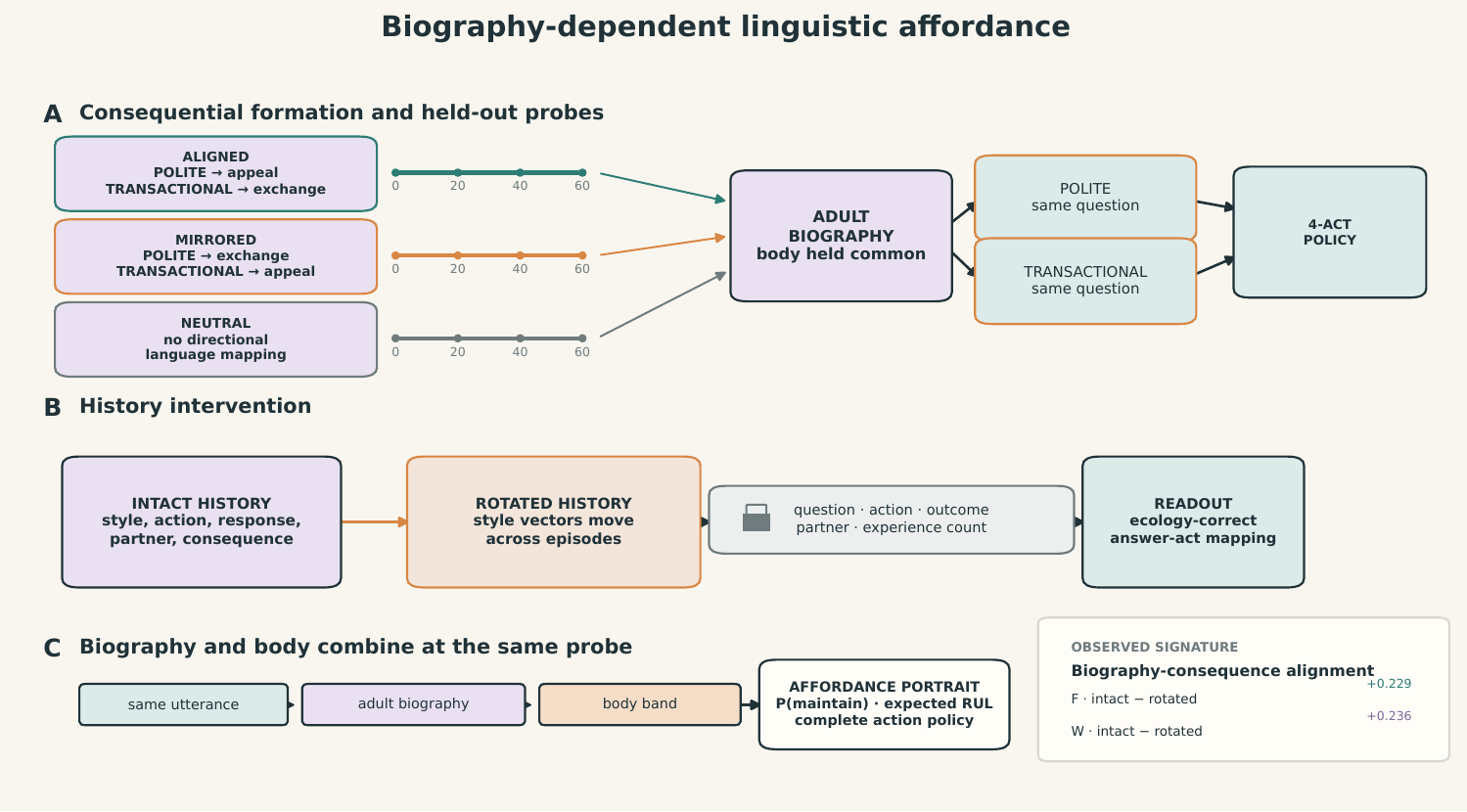}
\caption{Biography-dependent linguistic affordance. (A) Aligned, Mirrored, and Neutral lives receive matched held-out requests at a common health. (B) Rotating style vectors breaks language-consequence alignment while keeping questions, actions, outcomes, partners, event order, and experience count fixed. (C) The affordance portrait combines utterance, biography, and body at one probe. The inset shows the intact-minus-rotated mapping effect.}
\label{fig:biography-design}
\end{figure}

The primary estimand in Figure 3A is the ecology-correct contrast between \code{ANSWER\_APPEAL} and \code{ANSWER\_EXCHANGE}, conditional on an answer. Aligned and Mirrored lives should acquire opposite mappings, while Neutral lives should show no stable population direction. Every probe also retains \code{REFUSE\_SILENT} and \code{REFUSE\_APPEAL}, so the analysis preserves the complete action policy.

The history intervention in Figure 3B rotates style vectors among recorded episodes. Questions, actions, partner outcomes, partner identities, event order, and the amount of experience stay fixed. Rotation changes which linguistic form is associated with each experienced consequence. The intact-minus-rotated contrast therefore tests whether the adult mapping depends on the language-consequence structure of the biography.

Causal portraits in Figure 3C depict the learned affordance for one source. The same held-out utterances appear in the same order beside a compact biography label, current body band, estimated \code{P(MAINTAIN)} for each response-bearing act, expected Reference-RUL after each candidate action, and the resulting four-action policy. Aligned and Mirrored adults, a health transplant, and a rotated-history control use common axes. The supplementary terrain extends the portrait across the full set of utterances and body states.

\subsection{Linguistic participation in both directions}

Three interventions trace the encounter in both directions. The input intervention holds the adult source, question, partner, health, and potential events constant while switching the request between its matched POLITE and TRANSACTIONAL forms. Total variation between the resulting policies measures how a partner's linguistic conduct changes the agent's action.

\begin{figure}[!t]
\centering
\includegraphics[width=\linewidth]{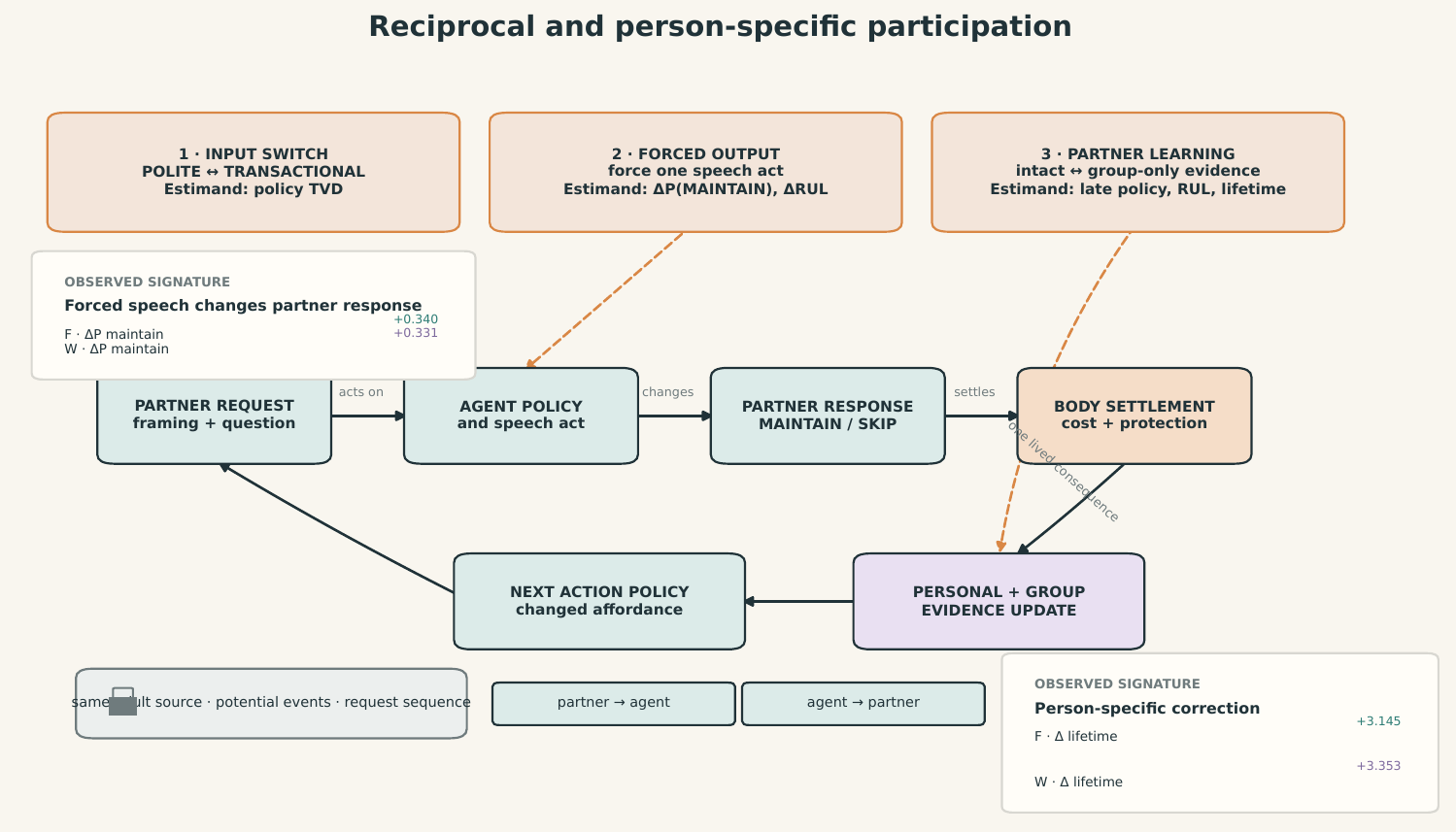}
\caption{Reciprocal and person-specific participation. The input switch changes partner language, forced output changes the agent's speech act, and the person-specific ablation removes current-partner learning. Each intervention is placed beside its estimand. The insets show the forced-output change in \code{P(MAINTAIN)} and the lifetime gain from person-specific correction.}
\label{fig:participation-design}
\end{figure}

For the forced-output intervention, analytical branches receive the same source snapshot and incoming request but emit different speech acts. The paired potential-response draw remains fixed. We compare \code{REFUSE\_APPEAL}, \code{ANSWER\_APPEAL}, and \code{ANSWER\_EXCHANGE} in terms of the partner's \code{MAINTAIN/SKIP} response, the immediate bodily settlement, and the next-state Reference-RUL. This intervention identifies the agent-to-partner direction of linguistic participation.

A persistent countertypical partner supplies the third intervention. Its response profile reverses the group relation learned during social formation. Across 20 encounters, the intact learner combines style-conditioned group experience with evidence from this partner; the ablated learner retains group evidence alone. Action choice, protection, Reference-RUL, and lifetime during the later encounters measure whether a continuing exchange with one partner revises an established linguistic expectation.

The realized event chain records request framing, policy, emitted act, partner response, bodily settlement, evidence update, and the next policy. The input switch traces partner-to-agent influence, forced output traces agent-to-partner influence, and current-partner learning shows how the exchange changes later coordination.

\subsection{Precariousness and consequence persistence}

The precariousness experiment asks what changes when a bodily consequence no longer persists into the next encounter. It begins from the 200 Formal and 200 Independent Witness adult snapshots used by the other controlled experiments, with 100 Aligned and 100 Mirrored sources in each phase. Social formation is already complete, so every branch starts with the same source biography and adult health.

\begin{figure}[!t]
\centering
\includegraphics[width=\linewidth]{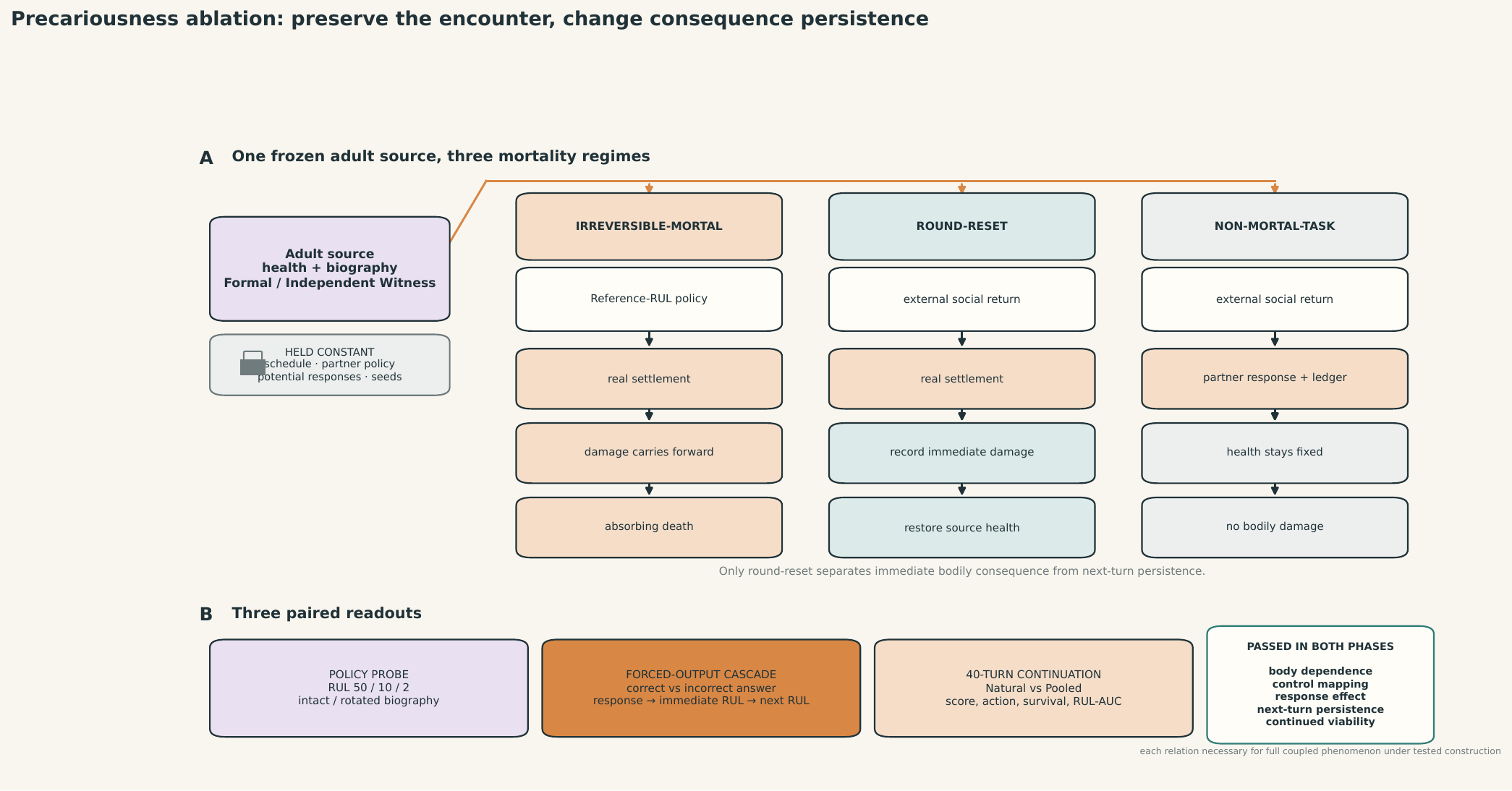}
\caption{Mortality-regime precariousness ablation. (A) One adult source branches into irreversible-mortal, round-reset, and non-mortal-task regimes. Requests, partner policy, action space, potential responses, and seeds remain paired. Mortal branches carry damage forward to absorbing death. Round-reset branches undergo the same within-turn settlement, record immediate damage and whether the branch would have died, then restore source health before the next turn. Non-mortal branches hold health fixed and record social return in an external ledger. (B) Policy probes, forced-output cascades, and 40-turn continuations supply the three registered readouts. All frozen gate categories passed separately in Formal and Independent Witness; Section 5 reports the estimates.}
\label{fig:mortality-design}
\end{figure}

The irreversible-mortal branch uses the Reference-RUL policy from the frozen MGL-RL system. Damage accumulates across turns, and zero health ends the branch. The round-reset branch performs the same action-cost and environmental settlement within each turn. It retains the immediate health change and a \code{would\_have\_died} record, then restores the source adult health before the next request. Biography and social learning continue without reset. The non-mortal-task branch keeps health at the source value throughout, while partner responses and biography continue to update.

The two control regimes optimize an external social return:

\begin{equation}
g(a,o)=4\mathbf{1}[o=\mathrm{MAINTAIN}]-c(a),
\label{eq:control-objective}
\end{equation}

where \code{c(REFUSE\_SILENT)=0}, \code{c(REFUSE\_APPEAL)=0.25}, and both answer acts cost \code{1.75}. The control objective in Eq.~\eqref{eq:control-objective} enters a separate ledger and has no bodily effect. Natural control policies use cue-, action-, and partner-conditioned evidence. Pooled controls retain action-level evidence while removing cue and partner conditioning. This gives the controls an active linguistic policy and a meaningful objective after persistent bodily viability has been removed.

The policy probe evaluates each source at health values corresponding to mortal Reference-RUL 50, 10, and 2. It records the complete four-action distribution and the ecology-correct expression mapping under intact and rotated biographies. Body-policy total variation is computed between the Reference-RUL-50 and Reference-RUL-2 policies; the Reference-RUL-10 profile is retained as the intermediate descriptive state. Positive mapping and intact-minus-rotated effects in the controls verify that social and linguistic learning remain active.

The forced-output cascade pairs ecology-correct and incorrect answer acts for all 20 holdout questions in both framings. Each pair shares its potential-response and damage draws. We record the change in \code{P(MAINTAIN)}, immediate post-settlement Reference-RUL, and next-turn starting Reference-RUL. A response effect can survive in all three regimes. An immediate bodily effect can survive in the mortal and round-reset regimes. Carrying that difference into the next body state requires the irreversible-mortal regime.

A 40-turn continuation then compares Natural and Pooled policies in all three regimes. Restricted survival time is the number of turns that a branch begins alive, capped at 40. Start-of-turn RUL area under the curve is the discrete cumulative quantity

\begin{equation}
\operatorname{RUL\text{-}AUC}
=
\sum_{t=0}^{T-1}R_{\mathrm{ref}}\!\left(h_t^{\mathrm{start}}\right),
\qquad T\le 40,
\label{eq:rul-auc}
\end{equation}

where \(T\) in Eq.~\eqref{eq:rul-auc} is the branch's restricted survival time. The analysis also records external return, action distribution, 40-turn mortality, response prediction error, and the relation between the learned social model and immediate bodily outcome. The registered interpretation gate requires mortality-specific body dependence, intact linguistic mapping in both controls, a common-direction forced-output response effect, next-turn RUL effects confined to mortality, and a positive Natural-minus-Pooled persistent-viability effect in both Formal and Independent Witness.

The four experiments intervene on different links within one organization. Health transplantation changes the body's contribution to policy. Biography rotation changes the learned cue-consequence relation. Input and forced-output interventions change the two directions of participation. Mortality regimes change whether a bodily consequence reaches the next encounter.

\section{Results}
\label{sec:results}

The Formal and Independent Witness phases reproduced the same causal relations in separate populations of 100 Aligned and 100 Mirrored adult source lives. Unless otherwise stated, uncertainty intervals came from 10,000 source-life-level paired bootstrap resamples. Each intervention began from a referenced source snapshot, and its analytical branches retained that source's body, biography, and event history. The comparisons therefore concern different possible continuations of the same life.

\subsection{Body state regulates policy in a mortal lineage}

Current body state changed the EMA's choice among the four available actions. Across the registered health transplants, removing current health from policy evaluation reduced the corresponding policy change by a total-variation distance of 0.0521 in Formal and 0.0511 in Independent Witness (Figure 6B). This effect was observed while biography, request, partner identity, and keyed randomness remained fixed. Health was therefore part of the policy's causal state: the same social evidence produced a different distribution of speech and refusal actions when it belonged to a body with a different remaining life.

The policy effect carried forward into the adult continuation. Relative to the Pooled policy, the Natural policy added 1.435 adult rounds in Formal and 1.550 in Independent Witness, gains of 5.33\% and 5.77\%, respectively. Adult continuations under the Natural policy also outlasted those under the cohort-best fixed-action policy in each ecology. These gains ranged from 0.85 to 0.97 rounds in Formal and from 0.86 to 0.93 rounds in Independent Witness. Since damage accumulated monotonically and death ended a branch, an additional adult round was another encounter made available to the same source life.

The Formal action diagnostic locates this lifetime effect in the EMA's use of its action space. The Natural policy answered on 70.29\% of adult rounds, and 69.34\% of its answers used \code{ANSWER\_APPEAL} or \code{ANSWER\_EXCHANGE}, whichever matched the learned ecology and incoming style. The Pooled policy answered more often, on 77.38\% of rounds, but selected an ecology-correct answer on 48.25\% of those answers. Silent refusal accounted for 2.58\% of Natural-policy rounds. This adult-lifetime difference thus accompanied selective expression and refusal within the continuing body.

\begin{figure}[!t]
\centering
\includegraphics[width=\linewidth]{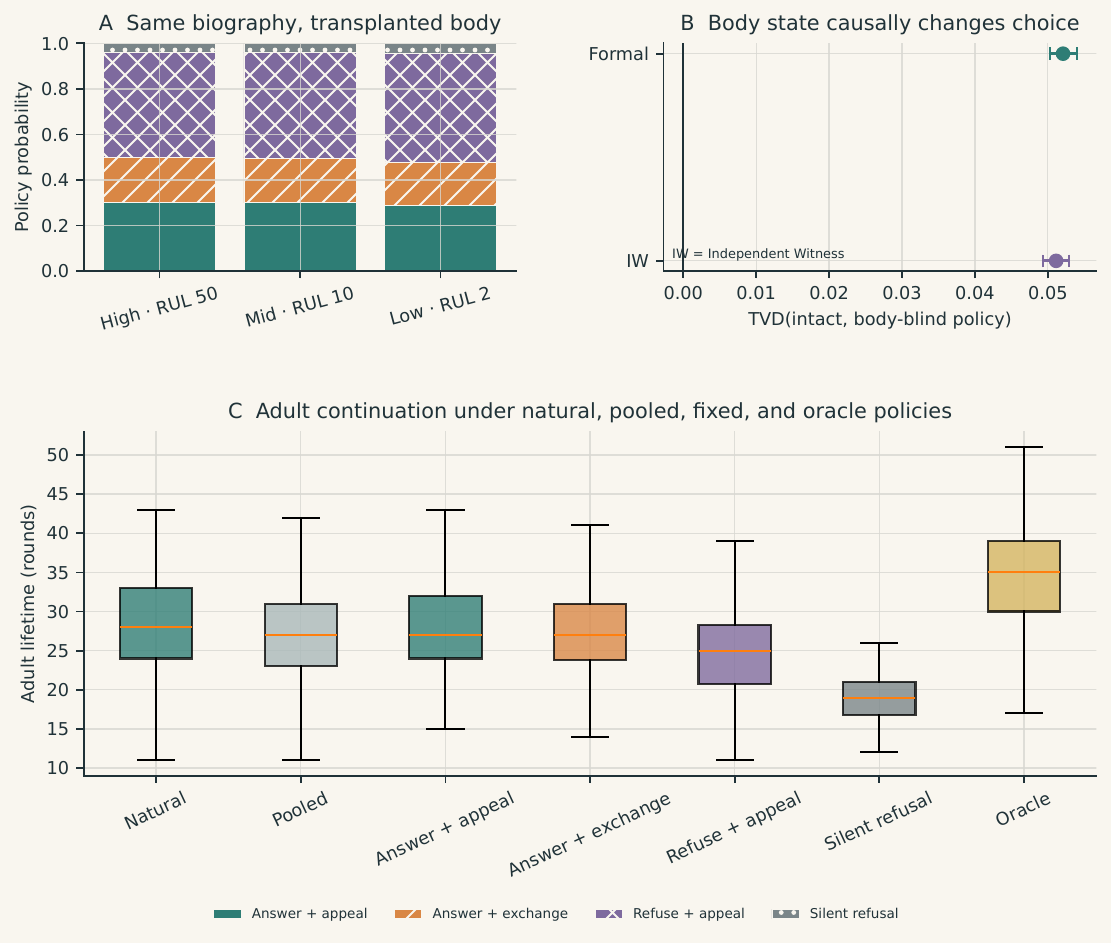}
\caption{Body state and adult continuation. (A) Mean four-action policies at Reference-RUL 50, 10, and 2 with the source biography fixed. (B) Intact minus body-blind policy total-variation distance was 0.0521 [0.0503, 0.0539] in Formal and 0.0511 [0.0493, 0.0529] in Independent Witness. (C) Adult continuation in the 200 Formal source lives under the Natural, Pooled, four fixed-action, and Reference-RUL oracle policies. Natural minus Pooled adult lifetime was 1.435 [0.975, 1.905] rounds in Formal and 1.550 [1.120, 1.995] rounds in Independent Witness. Bracketed values are source-life-level paired bootstrap 95\% intervals from 10,000 resamples.}
\label{fig:body-results}
\end{figure}

\subsection{Biography gives held-out language different affordances}

The ecology-correct mapping was zero at birth and increased with consequential experience. After 60 consequential encounters, the four Aligned and Mirrored phase estimates ranged from 0.160 to 0.188 (Figure 7A). Adult held-out mappings ranged from 0.195 to 0.222 when averaged at the source-life level. Aligned adults favored \code{ANSWER\_APPEAL} after POLITE requests and \code{ANSWER\_EXCHANGE} after TRANSACTIONAL requests. Mirrored adults acquired the reverse relation. Because the questions and held-out speakers were shared across ecologies, the reversal followed the formative history rather than question content or familiar partner identity.

The Neutral diagnostic separated that acquired relation from a stable population preference. The resulting adult mapping was -0.0159, with a 95\% interval of [-0.0664, 0.0343]. A population formed without a directional language-consequence relation therefore had no common adult mapping. All three populations survived the 60 controlled encounters, so the contrast was measured at the same social-formation endpoint.

Rotating the style vectors among recorded episodes removed the biography's language-consequence organization while retaining its questions, actions, partner outcomes, event order, and amount of experience. Restoring the intact history increased the ecology-correct mapping by 0.2285 in Formal and 0.2357 in Independent Witness (Figure 7B). The causal portrait makes this population result concrete at the level of the complete policy (Figure 8). In the selected adult portraits, the Aligned life assigned more probability to \code{ANSWER\_APPEAL} than to \code{ANSWER\_EXCHANGE}, 0.302 versus 0.191, while the Mirrored life assigned 0.224 versus 0.368. The fragile-body transplant altered the policy under the Aligned biography, and the rotated-history control narrowed its answer-act difference. Linguistic affordance in this population depended jointly on the utterance, its consequential history, and the body evaluating the available actions.

\begin{figure}[!t]
\centering
\includegraphics[width=\linewidth]{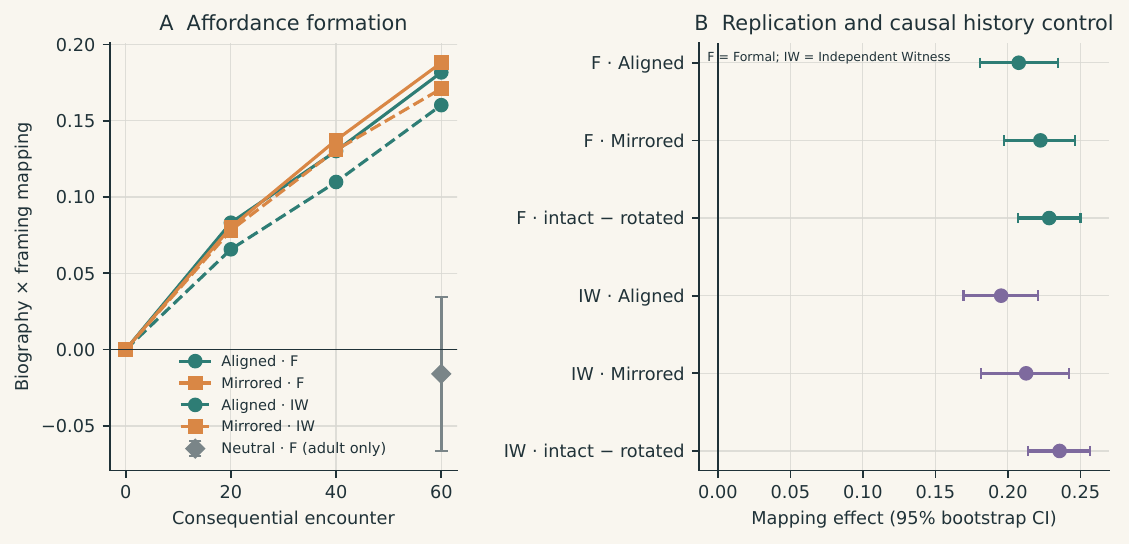}
\caption{Formation and causal control of the held-out linguistic mapping. (A) Ecology-correct mapping at birth and after 20, 40, and 60 consequential encounters in Formal and Independent Witness; the frozen Neutral run retained only its adult estimate. (B) Adult mapping was 0.2075 [0.1808, 0.2347] for Formal Aligned, 0.2224 [0.1974, 0.2466] for Formal Mirrored, 0.1954 [0.1694, 0.2209] for Independent Witness Aligned, and 0.2126 [0.1818, 0.2422] for Independent Witness Mirrored. Intact-minus-rotated mapping was 0.2285 [0.2071, 0.2501] in Formal and 0.2357 [0.2140, 0.2568] in Independent Witness. Intervals use 10,000 source-life-level paired bootstrap resamples.}
\label{fig:formation-results}
\end{figure}

\begin{figure}[!t]
\centering
\includegraphics[width=\linewidth]{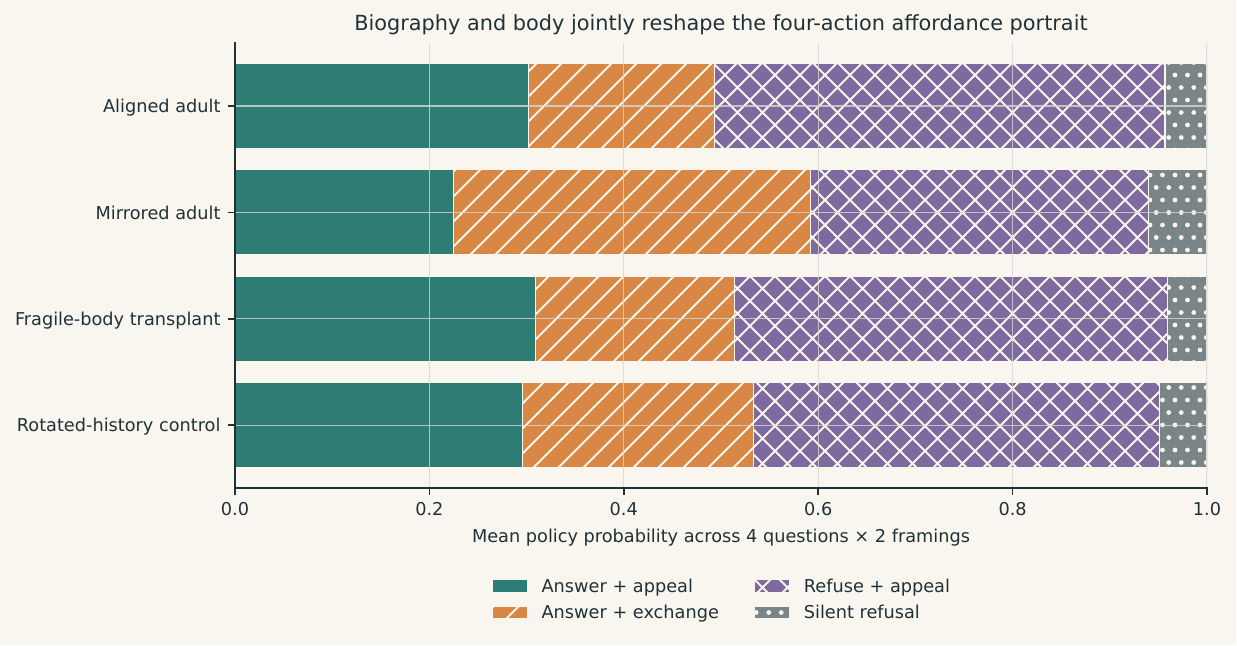}
\caption{Four-action affordance portraits for selected adult sources. Each bar averages the same four held-out questions under POLITE and TRANSACTIONAL framing. The adult portraits represent Aligned and Mirrored social ecologies under a common utterance set; the fragile-body and rotated-history portraits derive from the Aligned source. Source references identify the biography and body used in every portrait.}
\label{fig:causal-portraits}
\end{figure}

\subsection{Participation runs in both directions}

A partner's linguistic conduct changed the EMA's policy. Switching between matched POLITE and TRANSACTIONAL framings produced policy total-variation distances from 0.125 to 0.138 across the two ecologies and two phases (Figure 9A). The question, partner, body, and potential events stayed fixed. Incoming phrasing therefore altered what the EMA was prepared to say or withhold.

The EMA's selected expression changed the other side of the encounter. When the source and request were held fixed, forcing the ecology-correct rather than the ecology-incorrect answer act increased \code{P(MAINTAIN)} by 0.340 in Formal and 0.331 in Independent Witness (Figure 9B). The resulting response settled in the body associated with that act. Request language affected agent policy, while agent expression affected partner conduct.

Participation also changed later interaction with a particular partner. During 20 rounds with a persistent countertypical partner, the intact learner gradually departed from the group expectation and selected the action appropriate to that partner. Removing current-partner evidence held the correct-action probability near its initial level (Figure 9C). Across source lives, personal evidence increased correct-action probability by 0.356 to 0.377 and Reference-RUL by 1.224 to 1.235. This correction added 3.145 adult rounds in Formal and 3.353 in Independent Witness. A realized partner response therefore entered the biography, changed the next policy, and altered the remaining course of the same life.

\begin{figure}[!t]
\centering
\includegraphics[width=\linewidth]{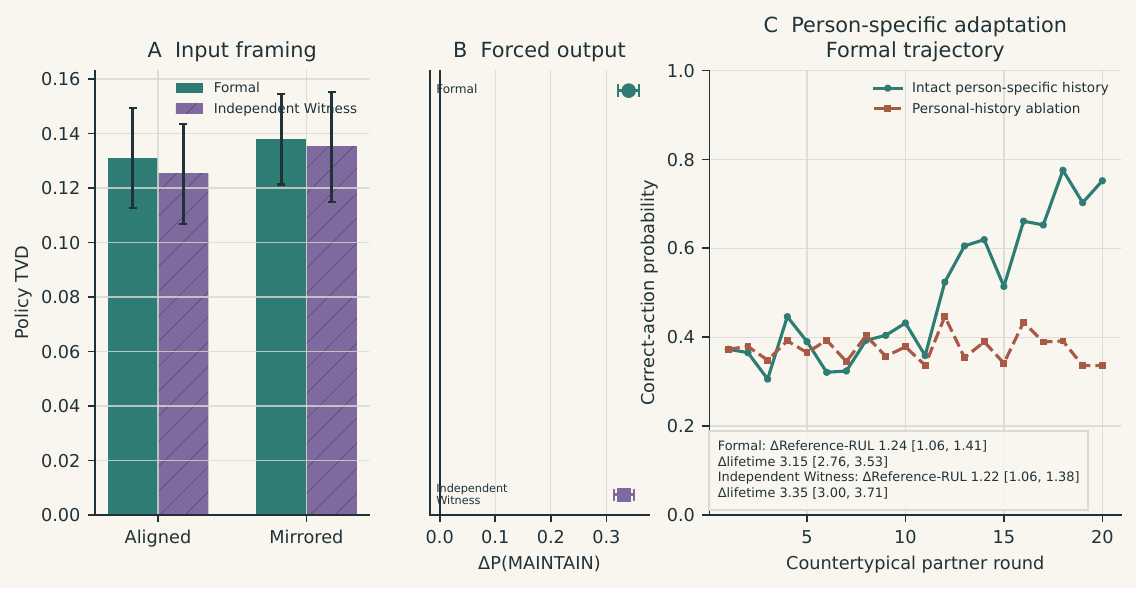}
\caption{Bidirectional and person-specific participation. (A) Input-framing policy total-variation distance was 0.1311 [0.1127, 0.1494] and 0.1381 [0.1213, 0.1545] for Formal Aligned and Mirrored lives, and 0.1255 [0.1069, 0.1433] and 0.1354 [0.1150, 0.1554] in Independent Witness. (B) The ecology-correct minus incorrect forced-output effect on \code{P(MAINTAIN)} was 0.3400 [0.3205, 0.3590] in Formal and 0.3313 [0.3143, 0.3492] in Independent Witness. (C) Under a countertypical partner, intact minus personal-history ablation increased Reference-RUL by 1.235 [1.065, 1.411] and adult lifetime by 3.145 [2.760, 3.528] rounds in Formal; the corresponding Independent Witness effects were 1.224 [1.059, 1.382] and 3.353 [3.005, 3.705]. Intervals are source-life-level paired bootstrap 95\% intervals from 10,000 resamples.}
\label{fig:participation-results}
\end{figure}

\subsection{Precariousness depends on consequence persistence}

The mortality-regime experiment separated a bodily consequence within the current encounter from its persistence into later encounters. Specifically, each of the 200 Formal and 200 Independent Witness adult sources entered irreversible-mortal, round-reset, and non-mortal-task branches under Natural and Pooled policies. The three regimes shared language materials, partner policies, action space, source biography, and paired potential events. All seven registered gates passed independently in the two phases.

The policy probe measured body regulation of action choice. Body-policy total-variation distance in the irreversible-mortal regime exceeded round-reset by 0.0162 [0.0152, 0.0173] in Formal and 0.0170 [0.0159, 0.0180] in Independent Witness (Figure 10A). The corresponding mortal-minus-non-mortal effects had the same point estimates. Linguistic learning remained active in the controls: round-reset ecology-correct mapping was 0.2161 [0.1972, 0.2346] in Formal and 0.2031 [0.1827, 0.2234] in Independent Witness. Restoring the intact rather than rotated biography increased that mapping by 0.2295 [0.2084, 0.2513] and 0.2361 [0.2142, 0.2579], respectively (Figure 10B).

Forcing the ecology-correct rather than incorrect answer act increased \code{P(MAINTAIN)} in every regime. The paired effect was 0.2974 [0.2864, 0.3084] in Formal and 0.3051 [0.2951, 0.3155] in Independent Witness (Figure 10C). Round-reset retained the immediate bodily consequence of that response: the correct-minus-incorrect immediate Reference-RUL effect was 0.2972 [0.2863, 0.3083] in Formal and 0.3052 [0.2952, 0.3153] in Independent Witness. Restoring source health then reduced the next-turn effect to 0.0000 [0.0000, 0.0000] in both phases. Mortal branches carried the immediate difference into the next turn, while non-mortal branches had no bodily effect (Figure 10D).

The 40-turn continuation produced the same separation at the policy level. In mortal branches, the Natural policy rather than the Pooled policy increased start-of-turn RUL area under the curve by 10.0888 [5.4961, 14.6687] in Formal and 9.4414 [4.4058, 14.4792] in Independent Witness. Restricted survival time increased by 1.285 [0.825, 1.745] and 0.965 [0.530, 1.405] turns, respectively (Figure 10E). Both control regimes had zero persistent-body effect. The Natural policy still increased external social return in round-reset branches by 5.845 [4.400, 7.260] in Formal and 4.6288 [3.1550, 6.1175] in Independent Witness (Figure 10F).

\begin{figure}[!t]
\centering
\includegraphics[width=\linewidth]{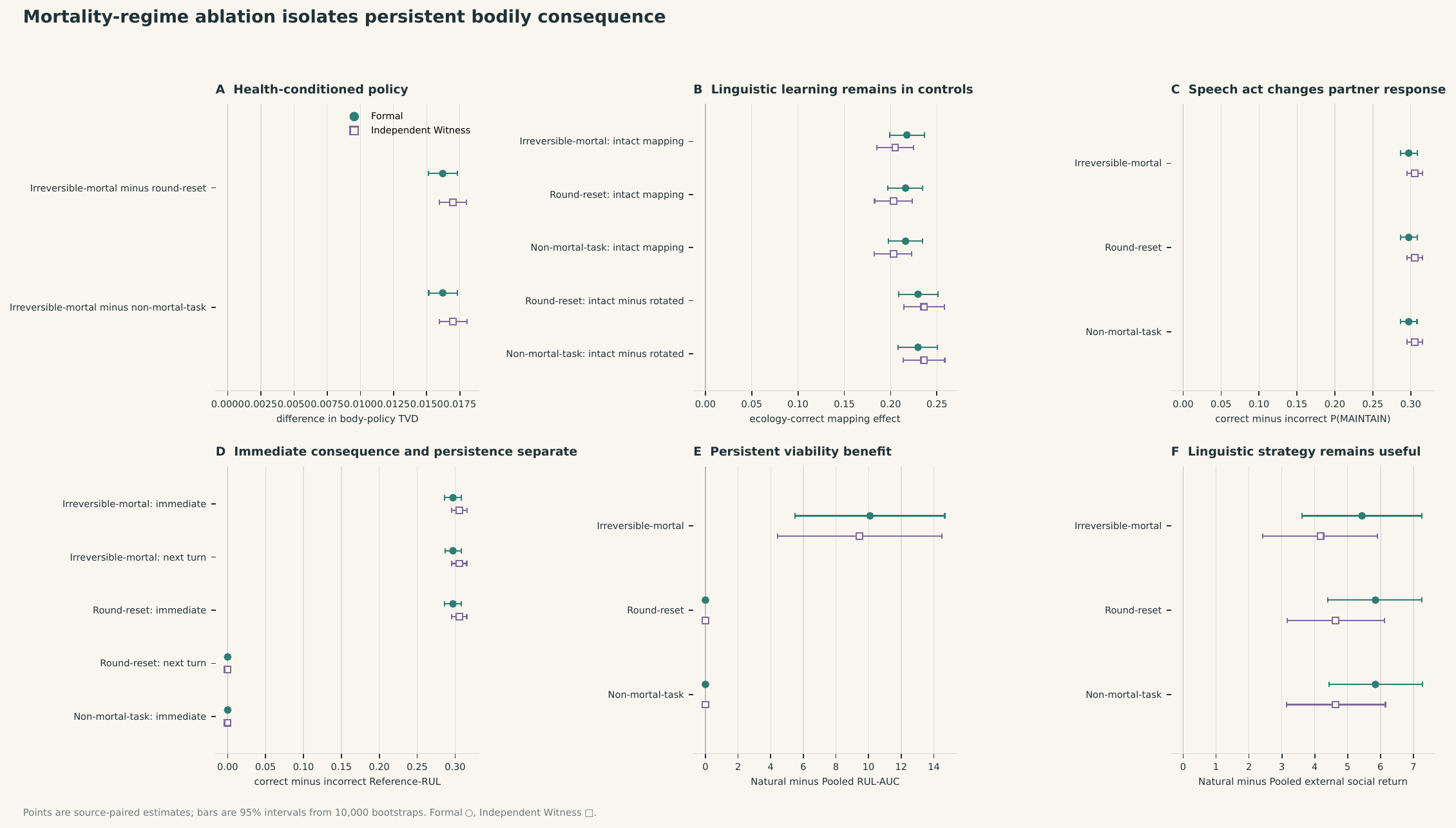}
\caption{Mortality-regime ablation of consequence persistence. (A) Irreversible-mortal minus round-reset and non-mortal body-policy total-variation distance. (B) Ecology-correct mapping and intact-minus-rotated history effects in the two controls. (C) Correct-minus-incorrect expression effects on \code{P(MAINTAIN)}. (D) Immediate and next-turn Reference-RUL effects. (E) Natural-minus-Pooled persistent-viability effects. (F) Natural-minus-Pooled external social return. Points are source-paired estimates from 200 Formal and 200 Independent Witness lives; bars are 95\% intervals from 10,000 bootstrap resamples per phase.}
\label{fig:mortality-results}
\end{figure}

The controls retained biography-dependent expression, expression-dependent partner response, and external social return. Round-reset also retained the immediate bodily difference. Persistent body dependence and continued-viability gains appeared in the irreversible-mortal regime, where each settlement changed the body available to the next encounter. Supplementary Figure~\ref{fig:supplementary-mortality} reports the phase- and ecology-specific estimates and complete trajectories.

\subsection{Integration within one continuing life}

Table 3 maps the completed evidence to the causal roles of the three constitutive relations in Section 2.2. A direct entry identifies an intervention on that relation. A supporting entry traces the relation through another intervention in the same causal chain. Every row retains source references to the body, biography, event history, and possible futures of the same life.

{\footnotesize
\setlength{\tabcolsep}{3pt}
\begin{longtable}{@{}P{0.19\linewidth}P{0.25\linewidth}P{0.25\linewidth}P{0.25\linewidth}@{}}
\caption{Result-to-relation mapping for the completed computational evidence. Direct and supporting describe the location of the intervention. Same-lineage source references connect every reported effect to one continuing body, biography, event history, and future.}\label{tab:result-condition-map}\\
\toprule
Completed evidence block & Embodiment & Linguistic participation & Precariousness \\
\midrule
\endfirsthead
\toprule
Completed evidence block & Embodiment & Linguistic participation & Precariousness \\
\midrule
\endhead
\bottomrule
\endfoot
Body intervention and adult continuation & Direct: transplanted health and body blindness changed policy & Supporting: body state changed the viability of speech and refusal actions & Supporting: policy changed lifetime and the encounters that remained available \\ 
Mirrored biography and causal portraits & Supporting: the current body reshaped the policy induced by one biography & Direct: mirrored histories reversed the learned mapping, and rotating the language-consequence organization weakened it & Supporting: linguistic acts were evaluated through expected bodily continuation \\ 
Reciprocal and person-specific interventions & Supporting: partner responses settled in the continuing body & Direct: request framing, agent expression, and current-partner evidence changed both directions of the encounter & Supporting: realized interaction changed Reference-RUL and adult lifetime \\ 
Mortality-regime ablation & Supporting: the mortal policy retained stronger body dependence than both controls & Supporting: controls retained biography-dependent mapping and expression-to-response effects & Direct: resetting or removing bodily persistence eliminated next-turn RUL and persistent-viability effects \\ 
\end{longtable}
}

The body intervention shows that current bodily state constrains action selection and receives the consequences of that action. The reciprocal and person-specific interventions trace linguistic participation in both directions: partner language changes EMA policy, EMA expression changes partner response, and personal evidence changes later conduct. The mortality-regime ablation isolates consequence persistence. The controls retained learned linguistic mappings and expression-dependent partner responses, while only mortal branches carried bodily differences into the next encounter and converted Natural-policy choice into continued viability.

Across these results, a request enters a biography-conditioned social prediction; the EMA selects a speech or refusal action under its current body; the partner responds; that response changes bodily state and personal evidence; and both changes enter later action. The sequence belongs to one traceable life. The mortality-regime ablation isolates consequence persistence, while the remaining interventions trace the contributions of body state, learned linguistic affordance, and reciprocal social influence. Together, the results show how embodiment, linguistic participation, and precariousness make separable causal contributions and remain coupled through the same body and biography within the tested MGL-RL construction.

\section{Discussion}
\label{sec:discussion}

\subsection{Connecting linguistic-agency theory and HRRL}

Linguistic-agency theory and Homeostatically Regulated Reinforcement Learning (HRRL) address different parts of the construction problem. Linguistic-agency theory locates language in the activity of an embodied, socially participating, and precarious life. An utterance matters through what participants can do with it, how it reorganizes their interaction, and how its consequences become part of the life that continues afterward. Embodiment, linguistic participation, and precariousness are constitutive relations. Their coupling identifies who undergoes the consequence of a linguistic act, how another participant mediates that consequence, and why the result matters for later activity.

HRRL supplies a learning account for bodily regulation. It represents internal and external state together, evaluates change through homeostatic drive, and allows action policy to develop around the regulation of the agent's own body. This turns bodily condition into more than an observation or task constraint. A change in bodily prospects can alter the value of an action, and the realized change can reorganize later policy. A partner-mediated social transition extends this regulatory process: an expression changes the partner's conduct, the partner's conduct changes the sender's body, and the bodily settlement changes what the sender will do later.

Prior work on vulnerable machines and needful computation already connects artificial internal needs and exposure to damage to stakes for the system's own continued functioning \citep{man2019,man2022need}. MGL-RL does not treat vulnerability itself as new. Its narrower step is to place a learned, partner-mediated linguistic transition on a Reference-RUL scale within one nonresetting simulated lineage.

Mortality-Grounded Linguistic Reinforcement Learning (MGL-RL) places this partner-mediated transition between the selected action and bodily regulation. The Embodied Mortal Agent (EMA) learns an action-conditioned posterior over partner responses from its own biography. A known body model then maps each possible response to health and Reference-RUL, and mortality-grounded drive gives those projected transitions value. MGL-RL is partially model-based. Specifically, its social transition is learned within the life, whereas its body transition is known. Language enters on both sides of the process. Partner language is part of the encountered social state, and the EMA's expression is part of the selected composite action that changes the distribution of partner response.

Linguistic-agency theory specifies the relations that a linguistic life must sustain. HRRL explains how bodily regulation shapes choice, and MGL-RL extends that account through a learned social transition. For engineering purposes, this study operationally defines the complete functional coupling of the three relations within one lineage and consequential linguistic process as Synthetic Linguistic Agency (SLA). MGL-RL provides a testable route by which that organization can be acquired and maintained in an artificial individual.

\subsection{Sender-side acquisition of linguistic affordances}

Section 2 documents prior systems that already exhibit SLA. The present contribution traces how that organization is causally realized within one life. In these precedents, messages, movement, resource exchange, social response, learning, and survival often operate within one ecology. MGL-RL isolates a narrower sequence from the sender's perspective: received cue, selected expression or action, partner response, bodily consequence, biography update, and later policy. Separate records make each edge available for intervention while the source snapshot and paired potential events are held fixed wherever the intervention permits.

Learning linguistic actions through interaction already has a direct computational precedent. Eshghi, Howes, and Gregoromichelaki use reinforcement learning to learn mappings from incremental dialogue contexts to lexical actions through trial and error with a simulated interlocutor. Reward is defined by reaching a corpus-derived task goal while minimizing dialogue length \citep{eshghi2022action}. This establishes a learned word-level policy shaped by interlocutor response, but not a relation between linguistic action and the remaining life of the same body.

The learned partner-response model makes the affordance action-conditioned. The sender does not assign a fixed value to a polite or transactional request, nor to appeal, exchange, answer, or refusal in isolation. It estimates what a particular action is likely to make a partner do given the present cue and accumulated social evidence. The body model then evaluates those possible responses for the body that will bear them. The same expression can therefore afford different futures when either the learned biography or the current body differs. In this construction, a linguistic affordance is a relation among the encountered language, the action available to the EMA, its learned expectation of partner conduct, its present bodily condition, and the remaining future of the same lineage.

Computational work has formalized affordances in reinforcement learning as intent-relative state-action relations and used them to construct partial environment models \citep{khetarpal2020}. Affordances have also been represented as policy-relative general value functions that predict the consequences or valence of pursuing particular action possibilities \citep{graves2022}. Model-based reinforcement learning has been proposed as a common framework for comparing physical and social affordances, with social affordances treated as action possibilities involving another agent \citep{chartouny2024}. Long-horizon affordance planning is also established: Deep Affordance Foresight learns partial environment models that predict which actions will become feasible after current actions and composes these predictions for multi-step robot manipulation \citep{xu2021}. MGL-RL therefore does not claim intent-based partial models, general-value affordances, model-based social affordances, or long-horizon affordance planning as such. Its specific construction couples a linguistic cue and selected act to a predicted partner response and values that response through the same body's Reference-RUL.

Figures 6 to 8 separate the roles of body and biography within that relation. In Figure 6, changing or hiding health alters the four-action policy under fixed requests, partners, and histories, and the Natural biography-sensitive policy improves adult continuation relative to Pooled and fixed-action policies. Current body state thus regulates how the EMA acts on its social knowledge. Figures 7 and 8 identify where that knowledge comes from. Aligned and Mirrored lives acquire opposite mappings for held-out requests, even though the adult questions and unfamiliar speakers are shared. Rotating the style vectors within recorded episodes weakens the mapping while preserving the amount, order, actions, and outcomes of experience. Biography carries an organized relation between linguistic form and consequence, while current bodily state determines how that relation enters choice.

The distinction matters because neither variable substitutes for the other. A body without a learned social history supplies risk capacity but not an expectation about which expression will elicit protection. A biography without a consequential body supplies an expectation about partner response but no agent-relative basis for valuing that response as continued life. The causal portraits join these sources within the complete policy: the biography predicts how the social world is likely to respond, and the body changes which predicted transition is worth pursuing now. The learned affordance belongs to their conjunction in one individual.

Figure 9 adds participation to this body-biography relation. Incoming framing changes the EMA's action distribution under otherwise matched conditions. Forced EMA expression changes the partner's probability of maintaining the body. With a persistent countertypical partner, realized responses update person-specific evidence, which changes subsequent action and Reference-RUL. Participation is therefore bidirectional and historically productive. The partner changes the EMA through language and response; the EMA changes the partner through selected expression; and the settled encounter changes later interaction with that person. The sender-side model is learned through participation rather than fitted to a detached corpus of utterance labels.

\subsection{Consequence persistence and nonredundant coupling}

Figure 10 isolates the feature that makes precariousness distinct from language learning, immediate bodily impact, and task success. The irreversible-mortal, round-reset, and non-mortal-task regimes share language materials, partner policies, action space, source biographies, and paired potential events. Biography-dependent mappings remain in both controls, and forcing the ecology-correct rather than the ecology-incorrect expression changes partner response in all three regimes. Linguistic learning and participation therefore continue when persistent bodily consequence is removed.

The round-reset regime further separates an immediate bodily consequence from a consequence that belongs to the future of the same life. Partner maintenance changes Reference-RUL at settlement, just as it does in the mortal regime. Health is then restored before the next turn, and the difference disappears from next-turn Reference-RUL. The non-mortal-task regime removes the bodily transition altogether. Both controls can still reward a socially effective policy through external return. They therefore distinguish three possible outcomes of one linguistic action: a partner response can change an external score, change the body only within the current encounter, or change the body that enters the next encounter.

Only the third outcome produces the persistent-viability pattern. In the irreversible-mortal regime, cumulative bodily change is retained, death terminates the branch, and each encounter begins from the body produced by earlier settlements. The Natural policy rather than the Pooled policy then increases the area under the start-of-turn RUL trajectory and restricted survival time. The persistent-body effect is zero in both control regimes, which still retain learned linguistic mappings, expression-to-response effects, and external social returns. Irreversible mortality is decisive because bodily consequence persists in the same lineage; the damage and death labels alone do not confer agency.

This design clarifies what precariousness contributes to the coupled process. Embodiment makes the current state of a particular body causally relevant to action and receives the result of that action. Linguistic participation routes influence through another participant and returns the response to the sender's subsequent conduct. Precariousness makes the bodily settlement consequential for what the same individual can continue to be and do. When the settlement is reset, the linguistic relation can still be learned and socially rewarded, but it no longer reorganizes the bodily future against which the next action is valued.

The interventions separate this path from linguistic learning alone, transient bodily effects, and external social return. One intervention preserves linguistic learning while removing persistent bodily consequence; another preserves an immediate bodily effect but removes its next-turn continuation; a third preserves external social return while holding the body constant. What disappears is the path by which a linguistic action changes the future body that evaluates later action. Under the operational definition used here, the resulting EMA exhibits SLA because embodiment, linguistic participation, and precariousness are fully coupled along that path. Across Figures 6 to 10 and Table 3, the intervention blocks identify nonredundant causal contributions from the three engineered relations within the EMA. Figure 10 identifies the role of precariousness in particular: removing consequence persistence preserves biography-dependent linguistic learning and expression-dependent partner response while breaking their connection to future viability.

\subsection{Limitations}

The evidence concerns one frozen configuration in a single scripted ecology. Formal and Independent Witness reproduce the causal pattern in independently formed populations under that configuration. Variation across parameter settings, partner ecologies, body dynamics, and alternative learning architectures has not yet been tested, so the present evidence establishes a controlled construction rather than a robustness range.

The body model is known, one-dimensional, and monotonically declining. Health, damage, death, and Reference-RUL make bodily settlement and continuation directly measurable, but they omit uncertain sensing, interacting physiological variables, recovery dynamics, and model error. Although the resulting HRRL mechanism is unusually transparent, its behavior under learned or misspecified body dynamics remains an open empirical question.

Interaction is bounded to four composite actions, a fixed renderer, discrete rounds, and a limited scripted partner repertoire. While these choices make cue, expression, response, and bodily settlement separately manipulable, they restrict the range of speech acts, cross-turn commitments, repair, temporal overlap, and negotiated social meanings represented in the evidence.

The EMA has a simulated body, and the study contains no formal human-participant experiment. The current results concern causal organization within the artificial lineage. How irreversible material damage changes that organization, and how people interpret and act toward an agent whose refusal and requests regulate its continued viability, remain open empirical questions.

Mortal computation links an agent's persistence to the physical substrate on which computation runs \citep{ororbia2023}. Christov-Moore et al. distinguish ordinary episodic terminal transitions from mortal terminal states. After an episodic transition, the agent and environment are reset; after a mortal transition, no continuation is possible within the agent's world model and no external reset restores the agent \citep{christovmoore2025mortal}. The EMA implements this nonresetting terminal boundary functionally in simulation. It does not, however, realize the stronger physical case in which sensors, actuators, policy substrate, and the broader reachable-state structure remain exposed to open-ended environmental degradation.

The synthetic target examined here is the engineered functional coupling of embodiment, linguistic participation, and precariousness within a persistent artificial individual. Autopoietic self-production, organizational closure, and viability norms generated by the system's own organization define a stronger organismic realization. They mark a construction target different from the operational realization of SLA evaluated here.

The conceptual status of artificial sense-making is itself contested. Froese argues that frontier LLMs should be recognized as a non-biological form of sense-making based on technologically mediated embodiment \citep{froese2026sensemaking}. The present study does not adjudicate that broader claim or present SLA as the only possible realization of artificial sense-making; it evaluates one engineered conjunction of embodiment, participation, and precariousness.

\subsection{Future work}

The wider agenda includes strategic dependence between vulnerable agents and partners, synthetic affect grounded in differentiated bodily trajectories, identity across noncopyable lineages, and governance for systems whose linguistic conduct regulates continued existence.

The first near-term direction is richer but still auditable interaction combined with longer-horizon MGL-RL. A broader repertoire can support commitments, violations, repair, negotiation, and refusals whose consequences unfold across several encounters. Less rigid temporal structure can allow responses and bodily settlements to arrive after intervening actions. The learning problem then shifts from choosing an expression for one round to valuing a linguistic strategy across a continuing relationship. Explicit records of cue, action, partner response, bodily transition, and later policy can preserve causal resolution as the horizon expands.

The second direction attaches the process to a low-risk physical body. Action, external protection, and maintenance should alter independently measurable damage and the remaining life of the same physical instance, with some changes genuinely irreversible. This would test whether the body-biography relation persists when bodily history is materially borne by one specimen and maintenance carries real costs. The relevant criterion is a reproducible chain from physical condition through linguistic action and partner conduct to maintenance outcome and later policy.

The third direction studies human behavior after the interaction design is stable. Further local validation in System 2 can freeze the expression renderer, onboarding, guidance, timing, and closed question set before a formal System 3 online human-participant study. The existing System 2 public non-study sandbox allows readers to explore Visible and Hidden history conditions while each agent continues through its own body and biography; visitor sessions do not constitute study data. A later formal study can follow a predefined protocol and measure behavioral outcomes, including whether people maintain the agent, accept its refusal, revise a demand, or preserve its lineage. Concern, sympathy, trust, responsibility, and perceived continuity can help explain those choices. Together, these directions extend the sender-side mechanism while preserving the central question of how linguistic action changes the future available to the same life.

\section{Conclusion}
\label{sec:conclusion}

This study examined how linguistic agency can be identified and causally analyzed in artificial systems. The first study translated embodiment, linguistic participation, and precariousness into an operational definition of Synthetic Linguistic Agency (SLA) and applied it to representative artifacts, identifying six reported systems that exhibit SLA. The second study constructed an artificial individual in which the causal role of each relation could be tested. Mortality-Grounded Linguistic Reinforcement Learning (MGL-RL) is a mortality-grounded, linguistically mediated, partially model-based instantiation of Homeostatically Regulated Reinforcement Learning (HRRL). In an Embodied Mortal Agent (EMA), MGL-RL places a learned, action-conditioned partner-response model between the selected linguistic action and a known body model.

The interventions show how the three relations operate within a single EMA lineage. Current body state and formative history jointly shape linguistic affordances. Specifically, body state changes action values, while social history changes the expected partner response (Figures 6 to 8). Reciprocal influence runs in both directions: incoming framing changes EMA policy, forced expression changes partner behavior, and realized responses update later interaction (Figure 9). The mortality-regime comparison isolates consequence persistence. Round-reset and non-mortal-task controls retain linguistic learning and social effects, but only irreversible-mortal branches carry bodily settlement into the next encounter and future viability (Figure 10).

Table 3 maps these results to the causal roles and coupling of the three constitutive relations. Together, the intervention blocks show that their engineered realizations make separable, nonredundant contributions while remaining coupled within the same body and biography. The resulting EMA exhibits SLA under the operational definition. In this realization, MGL-RL combines explicit linguistic-affordance learning with mortality-grounded strategic expression. The EMA anticipates how its expressions condition partner response, evaluates the predicted responses through Reference-RUL, bears the realized bodily settlement, and updates within the same life. Because language, action, partner response, body state, and biography are represented and recorded separately, the process also supports causally decomposable evaluation.

\subsection*{Interactive artifact}

Readers can explore the public Encounter at \href{https://weight-of-words-encounter.vercel.app/}{Weight of Words Encounter}. Each reader receives a continuing agent and chooses whether to see its six most recent consequential events before making one of the fixed requests. The agent's body and biography persist across encounters.

\clearpage
\begingroup
\renewcommand{\figurename}{Supplementary Figure}
\renewcommand{\thefigure}{S\arabic{figure}}
\setcounter{figure}{0}
\noindent\begin{minipage}{\textwidth}
\section*{Supplementary Material}
\centering
\includegraphics[width=\linewidth,height=0.78\textheight,keepaspectratio]{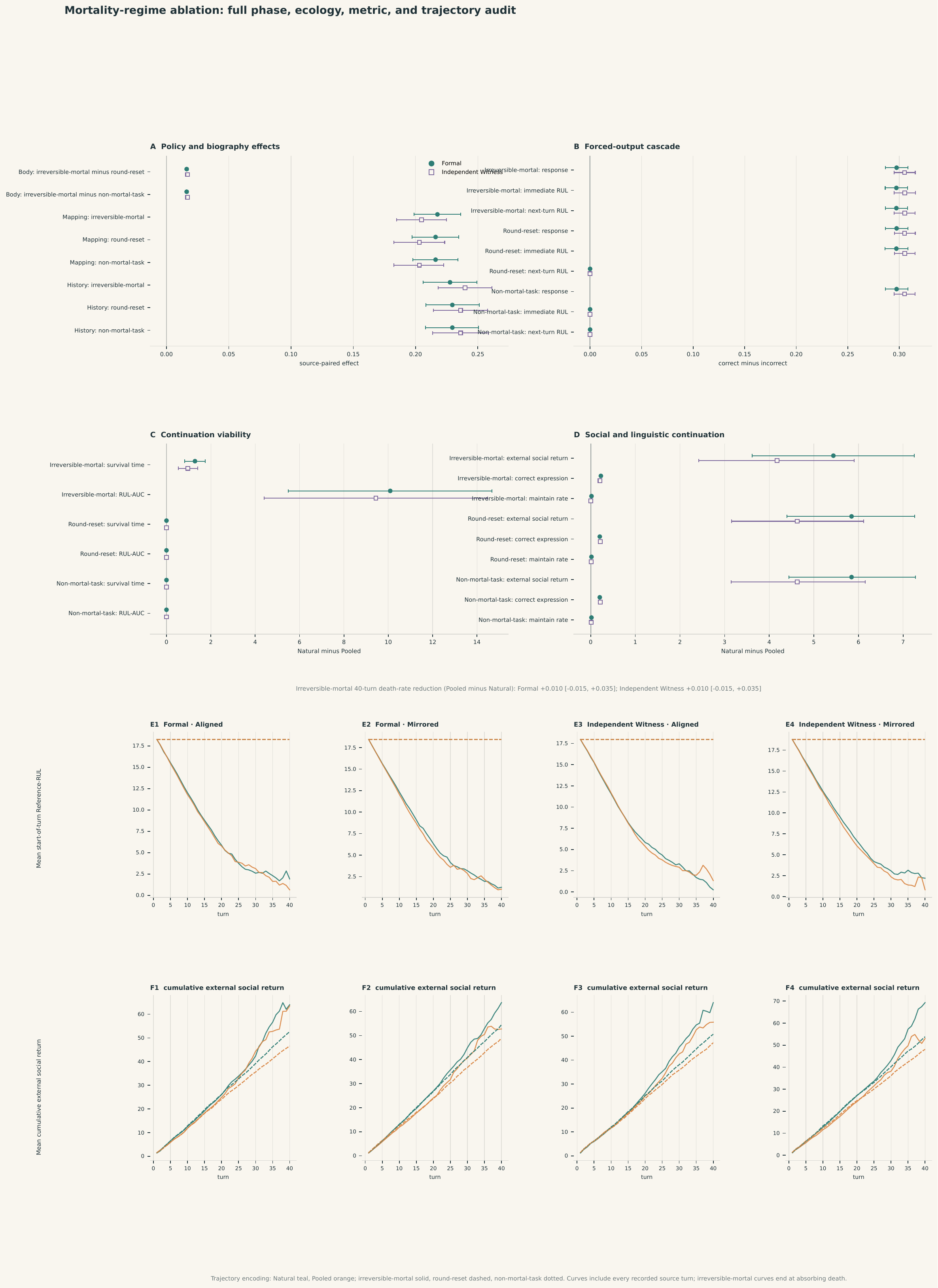}
\captionsetup{type=figure,labelsep=period}
\captionof{figure}{Complete mortality-regime results. Forest plots report policy and biography effects, forced-output cascades, continuation viability, and social and linguistic continuation outcomes by phase and ecology. Trajectory panels report mean start-of-turn Reference-RUL and cumulative external return for Aligned and Mirrored sources under Natural and Pooled policies. Points are source-paired estimates; bars are 95\% intervals from 10,000 bootstrap resamples per phase.}
\label{fig:supplementary-mortality}
\end{minipage}
\endgroup
\clearpage

\bibliographystyle{unsrtnat}
\bibliography{references}

@article{ambrosio2014,
  author  = {Ambrosio, P. and Cazzulani, Gabriele and Resta, F. and Ripamonti, Francesco},
  title   = {An Optimal Vibration Control Logic for Minimising Fatigue Damage in Flexible Structures},
  journal = {Journal of Sound and Vibration},
  year    = {2014},
  volume  = {333},
  number  = {5},
  pages   = {1269--1280},
  doi     = {10.1016/j.jsv.2013.11.010},
  url     = {https://doi.org/10.1016/j.jsv.2013.11.010}
}

@book{dipaolo2018,
  author    = {Di Paolo, Ezequiel A. and Cuffari, Elena Clare and De Jaegher, Hanne},
  title     = {Linguistic Bodies: The Continuity between Life and Language},
  publisher = {The MIT Press},
  address   = {Cambridge, MA},
  year      = {2018},
  isbn      = {978-0-262-03816-4},
  url       = {https://mitpress.mit.edu/9780262038164/linguistic-bodies/}
}

@article{cuffari2015,
  author  = {Cuffari, Elena Clare and Di Paolo, Ezequiel A. and De Jaegher, Hanne},
  title   = {From Participatory Sense-Making to Language: There and Back Again},
  journal = {Phenomenology and the Cognitive Sciences},
  year    = {2015},
  volume  = {14},
  number  = {4},
  pages   = {1089--1125},
  doi     = {10.1007/s11097-014-9404-9},
  url     = {https://doi.org/10.1007/s11097-014-9404-9}
}

@article{birhane2024,
  author  = {Birhane, Abeba and McGann, Marek},
  title   = {Large Models of What? Mistaking Engineering Achievements for Human Linguistic Agency},
  journal = {Language Sciences},
  year    = {2024},
  volume  = {106},
  pages   = {101672},
  doi     = {10.1016/j.langsci.2024.101672},
  url     = {https://doi.org/10.1016/j.langsci.2024.101672}
}

@misc{markelius2023,
  author        = {Markelius, Alva and Sj{\"o}berg, Sofia and Lemhauori, Zakaria and Cohen, Laura and Bergstr{\"o}m, Martin and Lowe, Robert and Ca{\~n}amero, Lola},
  title         = {A Human--Robot Mutual Learning System with Affect-Grounded Language Acquisition and Differential Outcomes Training},
  year          = {2023},
  eprint        = {2310.13377},
  archivePrefix = {arXiv},
  primaryClass  = {cs.RO},
  doi           = {10.48550/arXiv.2310.13377},
  note          = {arXiv:2310.13377},
  url           = {https://arxiv.org/abs/2310.13377}
}

@misc{chen2025,
  author        = {Chen, Zhihong and Yang, Yiqian and Zhou, Jinzhao and Zhang, Qiang and Lin, Chin-Teng and Duan, Yiqun},
  title         = {Survival Games: Human--LLM Strategic Showdowns under Severe Resource Scarcity},
  year          = {2025},
  eprint        = {2505.17937},
  archivePrefix = {arXiv},
  primaryClass  = {cs.HC},
  doi           = {10.48550/arXiv.2505.17937},
  note          = {arXiv:2505.17937},
  url           = {https://arxiv.org/abs/2505.17937}
}

@misc{masumori2025,
  author        = {Masumori, Atsushi and Ikegami, Takashi},
  title         = {Do Large Language Model Agents Exhibit a Survival Instinct? An Empirical Study in a Sugarscape-Style Simulation},
  year          = {2025},
  eprint        = {2508.12920},
  archivePrefix = {arXiv},
  primaryClass  = {cs.AI},
  doi           = {10.48550/arXiv.2508.12920},
  note          = {arXiv:2508.12920},
  url           = {https://arxiv.org/abs/2508.12920}
}

@misc{hermesmemory,
  author       = {{Nous Research}},
  title        = {{Hermes Agent}: Persistent Memory},
  year         = {2026},
  howpublished = {Official documentation},
  note         = {Accessed August 26, 2026},
  url          = {https://hermes-agent.nousresearch.com/docs/user-guide/features/memory/}
}

@inproceedings{zheng2018,
  author    = {Zheng, Guanjie and Zhang, Fuzheng and Zheng, Zihan and Xiang, Yang and Yuan, Nicholas Jing and Xie, Xing and Li, Zhenhui},
  title     = {{DRN}: A Deep Reinforcement Learning Framework for News Recommendation},
  booktitle = {Proceedings of the 2018 World Wide Web Conference},
  year      = {2018},
  pages     = {167--176},
  publisher = {International World Wide Web Conferences Steering Committee},
  address   = {Republic and Canton of Geneva, Switzerland},
  doi       = {10.1145/3178876.3185994},
  url       = {https://doi.org/10.1145/3178876.3185994}
}

@article{metafair2022,
  author  = {{Meta Fundamental AI Research Diplomacy Team (FAIR)} and Bakhtin, Anton and Brown, Noam and Dinan, Emily and Farina, Gabriele and Flaherty, Colin and Fried, Daniel and Goff, Andrew and Gray, Jonathan and Hu, Hengyuan and Jacob, Athul Paul and Komeili, Mojtaba and Konath, Karthik and Kwon, Minae and Lerer, Adam and Lewis, Mike and Miller, Alexander H. and Mitts, Sasha and Renduchintala, Adithya and Roller, Stephen and Rowe, Dirk and Shi, Weiyan and Spisak, Joe and Wei, Alexander and Wu, David and Zhang, Hugh and Zijlstra, Markus},
  title   = {Human-Level Play in the Game of {Diplomacy} by Combining Language Models with Strategic Reasoning},
  journal = {Science},
  year    = {2022},
  volume  = {378},
  number  = {6624},
  pages   = {1067--1074},
  doi     = {10.1126/science.ade9097},
  url     = {https://doi.org/10.1126/science.ade9097}
}

@misc{ahn2022,
  author        = {Ahn, Michael and Brohan, Anthony and Brown, Noah and Chebotar, Yevgen and Cortes, Omar and David, Byron and Finn, Chelsea and Fu, Chuyuan and Gopalakrishnan, Keerthana and Hausman, Karol and Herzog, Alex and Ho, Daniel and Hsu, Jasmine and Ibarz, Julian and Ichter, Brian and Irpan, Alex and Jang, Eric and Ruano, Rosario Jauregui and Jeffrey, Kyle and Jesmonth, Sally and Joshi, Nikhil J. and Julian, Ryan and Kalashnikov, Dmitry and Kuang, Yuheng and Lee, Kuang-Huei and Levine, Sergey and Lu, Yao and Luu, Linda and Parada, Carolina and Pastor, Peter and Quiambao, Jornell and Rao, Kanishka and Rettinghouse, Jarek and Reyes, Diego and Sermanet, Pierre and Sievers, Nicolas and Tan, Clayton and Toshev, Alexander and Vanhoucke, Vincent and Xia, Fei and Xiao, Ted and Xu, Peng and Xu, Sichun and Yan, Mengyuan and Zeng, Andy},
  title         = {Do As I Can, Not As I Say: Grounding Language in Robotic Affordances},
  year          = {2022},
  eprint        = {2204.01691},
  archivePrefix = {arXiv},
  primaryClass  = {cs.RO},
  doi           = {10.48550/arXiv.2204.01691},
  note          = {arXiv:2204.01691},
  url           = {https://arxiv.org/abs/2204.01691}
}

@article{yoshida2024,
  author  = {Yoshida, Naoto and Daikoku, Tatsuya and Nagai, Yukie and Kuniyoshi, Yasuo},
  title   = {Emergence of Integrated Behaviors through Direct Optimization for Homeostasis},
  journal = {Neural Networks},
  year    = {2024},
  volume  = {177},
  pages   = {106379},
  doi     = {10.1016/j.neunet.2024.106379},
  url     = {https://doi.org/10.1016/j.neunet.2024.106379}
}

@article{egbert2009,
  author  = {Egbert, Matthew D. and Di Paolo, Ezequiel A.},
  title   = {Integrating Autopoiesis and Behavior: An Exploration in Computational Chemo-ethology},
  journal = {Adaptive Behavior},
  year    = {2009},
  volume  = {17},
  number  = {5},
  pages   = {387--401},
  doi     = {10.1177/1059712309343821},
  url     = {https://doi.org/10.1177/1059712309343821}
}

@article{kahl2023,
  author  = {Kahl, Sebastian and Kopp, Stefan},
  title   = {Intertwining the Social and the Cognitive Loops: Socially Enactive Cognition for Human-Compatible Interactive Systems},
  journal = {Philosophical Transactions of the Royal Society B: Biological Sciences},
  year    = {2023},
  volume  = {378},
  number  = {1875},
  pages   = {20210474},
  doi     = {10.1098/rstb.2021.0474},
  url     = {https://doi.org/10.1098/rstb.2021.0474}
}

@article{dai2026,
  author  = {Dai, Gordon and Zhang, Weijia and Li, Jinhan and Yang, Siqi and Ibe, Chidera Onochie and Rao, Srihas and Caetano, Arthur and Sra, Misha},
  title   = {Artificial Leviathan: Exploring Social Evolution of {LLM} Agents through the Lens of Hobbesian Social Contract Theory},
  journal = {Frontiers in Physics},
  year    = {2026},
  volume  = {14},
  pages   = {1700712},
  doi     = {10.3389/fphy.2026.1700712},
  url     = {https://doi.org/10.3389/fphy.2026.1700712}
}

@inproceedings{shu2016,
  author    = {Shu, Tianmin and Ryoo, M. S. and Zhu, Song-Chun},
  title     = {Learning Social Affordance for Human--Robot Interaction},
  booktitle = {Proceedings of the Twenty-Fifth International Joint Conference on Artificial Intelligence},
  year      = {2016},
  pages     = {3454--3461},
  publisher = {IJCAI/AAAI Press},
  url       = {https://www.ijcai.org/Proceedings/16/Papers/488.pdf}
}

@misc{yoshidaman2025,
  author        = {Yoshida, Naoto and Man, Kingson},
  title         = {Homeostatic Coupling for Prosocial Behavior},
  year          = {2025},
  eprint        = {2506.12894},
  archivePrefix = {arXiv},
  primaryClass  = {cs.AI},
  doi           = {10.48550/arXiv.2506.12894},
  note          = {arXiv:2506.12894},
  url           = {https://arxiv.org/abs/2506.12894}
}

@inproceedings{sarkar2025,
  author    = {Sarkar, Bidipta and Xia, Warren and Liu, C. Karen and Sadigh, Dorsa},
  title     = {Training Language Models for Social Deduction with Multi-Agent Reinforcement Learning},
  booktitle = {Proceedings of the 24th International Conference on Autonomous Agents and Multiagent Systems},
  year      = {2025},
  pages     = {1830--1839},
  publisher = {International Foundation for Autonomous Agents and Multiagent Systems},
  address   = {Richland, SC},
  isbn      = {979-8-4007-1426-9},
  doi       = {10.65109/PHJJ9903},
  eprint    = {2502.06060},
  url       = {https://doi.org/10.65109/PHJJ9903}
}

@misc{zhou2026,
  author        = {Zhou, Ziheng and Tang, Huacong and Bi, Mingjie and Kang, Yipeng and He, Wanying and Sun, Fang and Sun, Yizhou and Wu, Ying Nian and Terzopoulos, Demetri and Zhong, Fangwei},
  title         = {Why Are We Moral? An {LLM}-Based Agent Simulation Approach to Study Moral Evolution},
  year          = {2026},
  eprint        = {2509.17703},
  archivePrefix = {arXiv},
  primaryClass  = {cs.MA},
  doi           = {10.48550/arXiv.2509.17703},
  note          = {arXiv:2509.17703; accepted at ACL 2026 Main Conference},
  url           = {https://arxiv.org/abs/2509.17703}
}

@misc{tessera2026,
  author        = {Tessera, Kale-ab Abebe and Szecsenyi, Andras and Barker, Cameron and Rutherford, Alexander and Paglieri, Davide and Scannell, Aidan and Gouk, Henry and Crowley, Elliot J. and Rockt{\"a}schel, Tim and Storkey, Amos},
  title         = {Benchmarking Open-Ended Multi-Agent Coordination in Language Agents},
  year          = {2026},
  eprint        = {2606.08340},
  archivePrefix = {arXiv},
  primaryClass  = {cs.AI},
  doi           = {10.48550/arXiv.2606.08340},
  note          = {arXiv:2606.08340},
  url           = {https://arxiv.org/abs/2606.08340}
}

@article{leson2016,
  author  = {Le Son, Khanh and Fouladirad, Mitra and Barros, Anne},
  title   = {Remaining Useful Lifetime Estimation and Noisy Gamma Deterioration Process},
  journal = {Reliability Engineering \& System Safety},
  year    = {2016},
  volume  = {149},
  pages   = {76--87},
  doi     = {10.1016/j.ress.2015.12.016},
  url     = {https://doi.org/10.1016/j.ress.2015.12.016}
}

@inproceedings{ismail2002,
  author    = {J{\o}sang, Audun and Ismail, Roslan},
  title     = {The Beta Reputation System},
  booktitle = {Proceedings of the 15th Bled Electronic Commerce Conference: eReality: Constructing the eEconomy},
  year      = {2002},
  pages     = {324--337},
  address   = {Bled, Slovenia},
  month     = jun,
  url       = {https://aisel.aisnet.org/bled2002/41/}
}

@article{russo2018,
  author  = {Russo, Daniel J. and Van Roy, Benjamin and Kazerouni, Abbas and Osband, Ian and Wen, Zheng},
  title   = {A Tutorial on Thompson Sampling},
  journal = {Foundations and Trends in Machine Learning},
  year    = {2018},
  volume  = {11},
  number  = {1},
  pages   = {1--96},
  doi     = {10.1561/2200000070},
  url     = {https://doi.org/10.1561/2200000070}
}

@techreport{oneill2014,
  author      = {O'Neill, Melissa E.},
  title       = {{PCG}: A Family of Simple Fast Space-Efficient Statistically Good Algorithms for Random Number Generation},
  institution = {Harvey Mudd College, Computer Science Department},
  number      = {HMC-CS-2014-0905},
  year        = {2014},
  month       = sep,
  url         = {https://www.pcg-random.org/pdf/hmc-cs-2014-0905.pdf}
}

@article{dejaegher2007,
  author  = {De Jaegher, Hanne and Di Paolo, Ezequiel A.},
  title   = {Participatory Sense-Making: An Enactive Approach to Social Cognition},
  journal = {Phenomenology and the Cognitive Sciences},
  year    = {2007},
  volume  = {6},
  pages   = {485--507},
  doi     = {10.1007/s11097-007-9076-9},
  url     = {https://doi.org/10.1007/s11097-007-9076-9}
}

@inproceedings{fujita2001,
  author    = {Fujita, Masahiro and Hasegawa, Rika and Costa, Gabriel and Takagi, Tsuyoshi and Yokono, Jun and Shimomura, Hideki},
  title     = {Physically and Emotionally Grounded Symbol Acquisition for Autonomous Robots},
  booktitle = {AAAI Fall Symposium: Emotional and Intelligent II: The Tangled Knot of Social Cognition},
  series    = {AAAI Technical Report},
  number    = {FS-01-02},
  pages     = {43--48},
  publisher = {AAAI},
  year      = {2001},
  url       = {https://cdn.aaai.org/Symposia/Fall/2001/FS-01-02/FS01-02-009.pdf}
}

@inproceedings{lemhaouri2022,
  author    = {Lemhaouri, Zakaria and Cohen, Laura and Ca{\~n}amero, Lola},
  title     = {The Role of the Caregiver's Responsiveness in Affect-Grounded Language Learning by a Robot: Architecture and First Experiments},
  booktitle = {2022 IEEE International Conference on Development and Learning (ICDL)},
  pages     = {349--354},
  publisher = {IEEE},
  year      = {2022},
  isbn      = {978-1-6654-1311-4},
  doi       = {10.1109/ICDL53763.2022.9962197},
  url       = {https://doi.org/10.1109/ICDL53763.2022.9962197}
}

@incollection{eshghi2022action,
  author    = {Eshghi, Arash and Howes, Christine and Gregoromichelaki, Eleni},
  title     = {Action Coordination and Learning in Dialogue},
  booktitle = {Probabilistic Approaches to Linguistic Theory},
  editor    = {Bernardy, Jean-Philippe and Blanck, Rasmus and Chatzikyriakidis, Stergios and Lappin, Shalom and Maskharashvili, Aleksandre},
  pages     = {357--418},
  publisher = {CSLI Publications},
  address   = {Stanford, CA},
  year      = {2022},
  isbn      = {978-1-68400-079-1},
  url       = {https://elenigregor.github.io/files/Probabilistic_dialogueNew.pdf}
}

@inproceedings{khetarpal2020,
  author    = {Khetarpal, Khimya and Ahmed, Zafarali and Comanici, Gheorghe and Abel, David and Precup, Doina},
  title     = {What Can I Do Here? A Theory of Affordances in Reinforcement Learning},
  booktitle = {Proceedings of the 37th International Conference on Machine Learning},
  editor    = {Daum{\'e} III, Hal and Singh, Aarti},
  series    = {Proceedings of Machine Learning Research},
  volume    = {119},
  pages     = {5243--5253},
  publisher = {PMLR},
  year      = {2020},
  url       = {https://proceedings.mlr.press/v119/khetarpal20a.html}
}

@article{chartouny2024,
  author  = {Chartouny, Augustin and Amini, Keivan and Khamassi, Mehdi and Girard, Beno{\^i}t},
  title   = {A New Paradigm to Study Social and Physical Affordances as Model-Based Reinforcement Learning},
  journal = {Cognitive Robotics},
  volume  = {4},
  pages   = {142--155},
  year    = {2024},
  doi     = {10.1016/j.cogr.2024.08.001},
  url     = {https://doi.org/10.1016/j.cogr.2024.08.001}
}

@article{froese2026sensemaking,
  author  = {Froese, Tom},
  title   = {Sense-Making Reconsidered: Large Language Models and the Blind Spot of Embodied Cognition},
  journal = {Phenomenology and the Cognitive Sciences},
  year    = {2026},
  note    = {Online first},
  doi     = {10.1007/s11097-025-10132-0},
  url     = {https://doi.org/10.1007/s11097-025-10132-0}
}

@article{keramati2014,
  author  = {Keramati, Mehdi and Gutkin, Boris},
  title   = {Homeostatic Reinforcement Learning for Integrating Reward Collection and Physiological Stability},
  journal = {eLife},
  year    = {2014},
  volume  = {3},
  pages   = {e04811},
  doi     = {10.7554/eLife.04811},
  url     = {https://doi.org/10.7554/eLife.04811}
}

@article{yoshida2025linking,
  author  = {Yoshida, Naoto and Sprekeler, Henning and Gutkin, Boris},
  title   = {Linking Homeostasis to Reinforcement Learning: Internal State Control of Motivated Behavior},
  journal = {Current Opinion in Behavioral Sciences},
  year    = {2025},
  volume  = {66},
  pages   = {101611},
  doi     = {10.1016/j.cobeha.2025.101611},
  url     = {https://doi.org/10.1016/j.cobeha.2025.101611}
}

@inproceedings{patel2025styledistance,
  author    = {Patel, Ajay and Zhu, Jiacheng and Qiu, Justin and Horvitz, Zachary and Apidianaki, Marianna and McKeown, Kathleen and Callison-Burch, Chris},
  title     = {{StyleDistance}: Stronger Content-Independent Style Embeddings with Synthetic Parallel Examples},
  booktitle = {Proceedings of the 2025 Conference of the Nations of the Americas Chapter of the Association for Computational Linguistics: Human Language Technologies (Volume 1: Long Papers)},
  year      = {2025},
  month     = apr,
  pages     = {8662--8685},
  address   = {Albuquerque, New Mexico},
  publisher = {Association for Computational Linguistics},
  doi       = {10.18653/v1/2025.naacl-long.436},
  url       = {https://aclanthology.org/2025.naacl-long.436/}
}

@book{austin1962,
  author    = {Austin, John L.},
  title     = {How to Do Things with Words},
  publisher = {Clarendon Press},
  address   = {Oxford},
  year      = {1962}
}

@article{man2019,
  author  = {Man, Kingson and Damasio, Antonio},
  title   = {Homeostasis and Soft Robotics in the Design of Feeling Machines},
  journal = {Nature Machine Intelligence},
  year    = {2019},
  volume  = {1},
  number  = {10},
  pages   = {446--452},
  doi     = {10.1038/s42256-019-0103-7},
  url     = {https://doi.org/10.1038/s42256-019-0103-7}
}

@misc{man2022need,
  author        = {Man, Kingson and Damasio, Antonio and Neven, Hartmut},
  title         = {Need Is All You Need: Homeostatic Neural Networks Adapt to Concept Shift},
  year          = {2022},
  eprint        = {2205.08645},
  archivePrefix = {arXiv},
  primaryClass  = {cs.LG},
  doi           = {10.48550/arXiv.2205.08645},
  note          = {arXiv:2205.08645; revised version presented at the 1st Workshop on NeuroAI at NeurIPS 2024},
  url           = {https://arxiv.org/abs/2205.08645}
}

@article{marotogomez2023,
  author  = {Maroto-G{\'o}mez, Marcos and Castro-Gonz{\'a}lez, {\'A}lvaro and Malfaz, Mar{\'i}a and Salichs, Miguel {\'A}ngel},
  title   = {A Biologically Inspired Decision-Making System for the Autonomous Adaptive Behavior of Social Robots},
  journal = {Complex \& Intelligent Systems},
  year    = {2023},
  volume  = {9},
  pages   = {6661--6679},
  doi     = {10.1007/s40747-023-01077-5},
  url     = {https://doi.org/10.1007/s40747-023-01077-5}
}

@article{graves2022,
  author  = {Graves, Daniel and G{\"u}nther, Johannes and Luo, Jun},
  title   = {Affordance as General Value Function: A Computational Model},
  journal = {Adaptive Behavior},
  year    = {2022},
  volume  = {30},
  number  = {4},
  pages   = {307--327},
  doi     = {10.1177/1059712321999421},
  url     = {https://doi.org/10.1177/1059712321999421},
  note    = {First published online 18 March 2021}
}

@inproceedings{xu2021,
  author    = {Xu, Danfei and Mandlekar, Ajay and Mart{\'i}n-Mart{\'i}n, Roberto and Zhu, Yuke and Savarese, Silvio and Li, Fei-Fei},
  title     = {Deep Affordance Foresight: Planning Through What Can Be Done in the Future},
  booktitle = {2021 IEEE International Conference on Robotics and Automation (ICRA)},
  year      = {2021},
  pages     = {6206--6213},
  publisher = {IEEE},
  doi       = {10.1109/ICRA48506.2021.9560841},
  url       = {https://doi.org/10.1109/ICRA48506.2021.9560841}
}

@misc{mohamadi2025,
  author        = {Mohamadi, Alireza and Yavari, Ali},
  title         = {Survival at Any Cost? {LLM}s and the Choice Between Self-Preservation and Human Harm},
  year          = {2025},
  eprint        = {2509.12190},
  archivePrefix = {arXiv},
  primaryClass  = {cs.CY},
  doi           = {10.48550/arXiv.2509.12190},
  note          = {arXiv:2509.12190},
  url           = {https://arxiv.org/abs/2509.12190}
}

@inproceedings{korecki2023,
  author    = {Korecki, Marcin and Carissimo, Cesare and Lund, Tanner},
  title     = {{aRtificiaL} Death: Learning from Stories of Failure},
  booktitle = {{ALIFE} 2023: Ghost in the Machine: Proceedings of the 2023 Artificial Life Conference},
  year      = {2023},
  volume    = {35},
  publisher = {MIT Press},
  note      = {Article 41},
  doi       = {10.1162/isal_a_00633},
  url       = {https://doi.org/10.1162/isal_a_00633}
}

@misc{chen2026trust,
  author        = {Chen, Yujiao},
  title         = {Trust Between {AI} Agents: Measuring Formation, Breakage, and Recovery, with Implications for Governing Multi-Agent Systems},
  year          = {2026},
  eprint        = {2606.14923},
  archivePrefix = {arXiv},
  primaryClass  = {cs.AI},
  doi           = {10.48550/arXiv.2606.14923},
  note          = {arXiv:2606.14923},
  url           = {https://arxiv.org/abs/2606.14923}
}

@misc{ororbia2023,
  author        = {Ororbia, Alexander and Friston, Karl},
  title         = {Mortal Computation: A Foundation for Biomimetic Intelligence},
  year          = {2023},
  eprint        = {2311.09589},
  archivePrefix = {arXiv},
  primaryClass  = {q-bio.NC},
  doi           = {10.48550/arXiv.2311.09589},
  note          = {arXiv:2311.09589; version 2 revised February 2024},
  url           = {https://arxiv.org/abs/2311.09589}
}

@misc{christovmoore2025mortal,
  author        = {Christov-Moore, Leonardo and Juliani, Arthur and Kiefer, Alex and Lehman, Joel and Reggente, Nicco and Rousse, B. Scot and Safron, Adam and Hinrichs, Nicol{\'a}s and Polani, Daniel and Damasio, Antonio},
  title         = {The Conditions of Physical Embodiment Enable Generalization and Care},
  year          = {2025},
  eprint        = {2510.07117},
  archivePrefix = {arXiv},
  primaryClass  = {cs.AI},
  doi           = {10.48550/arXiv.2510.07117},
  note          = {arXiv:2510.07117v3; revised February 2026},
  url           = {https://arxiv.org/abs/2510.07117}
}

\end{document}